\documentclass[final,onefignum,onetabnum]{siamonline171218}

\usepackage{lipsum}
\usepackage{amsfonts}
\usepackage{graphicx}
\usepackage{epstopdf}
\usepackage{algorithmic}
\ifpdf
  \DeclareGraphicsExtensions{.eps,.pdf,.png,.jpg}
\else
  \DeclareGraphicsExtensions{.eps}
\fi

\usepackage{enumitem}
\setlist[enumerate]{leftmargin=.5in}
\setlist[itemize]{leftmargin=.5in}

\newsiamremark{remark}{Remark}
\newsiamremark{hypothesis}{Hypothesis}
\crefname{hypothesis}{Hypothesis}{Hypotheses}
\newsiamthm{claim}{Claim}

\headers{Stochastic Training of ResNets}{Q. Sun, Y. Tao, and Q. Du}

\title{Stochastic Training of Residual Networks: a Differential Equation Viewpoint\thanks{Preprint. Work in progress.
\funding{The work of Qi Sun is partially supported by the NSFC program for Scientific Research Center under program No.: U1530401 and the China Scholarship Council No. 201606340009. The work of Qiang Du is partially supported by US National Science Foundation grant DMS-1719699, and NSF CCF-1740833.}}}

\author{Qi Sun\thanks{School of Mathematical Science, University of Science and Technology of China, Hefei 230026, China, and Division of Applied and Computational Mathematics, Beijing Computational Science Research Center, Beijing 100193, China
  (\email{sunqi@csrc.ac.cn}, \email{sunqi@mail.ustc.edu.cn}).}
\and Yunzhe Tao\thanks{Department of Applied Physics and Applied Mathematics, Columbia University, New York, NY 10027 
  (\email{y.tao@columbia.edu}).}
\and Qiang Du\thanks{Department of Applied Physics and Applied Mathematics and Data Science Institute, Columbia University, New York, NY 10027
  (\email{qd2125@columbia.edu}).}}

\usepackage{amsopn}

\usepackage{subcaption} 
\usepackage{tikz} 
\usetikzlibrary{arrows,shapes,positioning,automata}
\usetikzlibrary{calc,decorations.markings}
\usetikzlibrary{decorations.pathreplacing}
\usetikzlibrary{matrix}
\usepackage{subcaption} 
\usepackage{multirow,multicol,caption}
\usepackage{booktabs}

\ifpdf
\hypersetup{
  pdftitle={An Example Article},
  pdfauthor={D. Doe, P. T. Frank, and J. E. Smith}
}
\fi

\begin{document}

\maketitle


\begin{abstract}
During the last few years, significant attention has been paid to the stochastic training of artificial neural networks, which is known as an effective regularization approach that helps improve the generalization capability of trained models. In this work, the method of modified equations is applied to show that the residual network and its variants with noise injection can be regarded as weak approximations of stochastic differential equations. Such observations enable us to bridge the stochastic training processes with the optimal control of backward Kolmogorov's equations. This not only offers a novel perspective on the effects of regularization from the loss landscape viewpoint but also sheds light on the design of more reliable and efficient stochastic training strategies. As an example, we propose a new way to utilize Bernoulli dropout within the plain residual network architecture and conduct experiments on a real-world image classification task to substantiate our theoretical findings.
\end{abstract}

\begin{keywords}
Residual Networks, Stochastic Training Techniques, Stochastic Modified Equations, Optimal Control of Partial Differential Equations, Artificial Viscosity, Landscape Perspective
\end{keywords}

\begin{AMS}
49J20, 65C30, 62M45 
\end{AMS}

\section{Introduction}
Artificial neural networks are appearing as the state-of-the-art technologies in many machine learning tasks including, but not limited to, computer vision, speech recognition, and natural language processing \cite{lecun2015deep,schmidhuber2015deep}, among which the residual network (ResNet) and its numerous variants (see  \cite{he2016deep,gastaldi2017shake,huang2017densely,lu2017beyond,zagoruyko2016wide} and references cited therein) have attracted broad attention for their simplicity and effectiveness. In addition, ResNet makes it possible to train up to hundreds or even thousands of layers without performance degradation \cite{he2016identity}. However, deeper networks usually require extensive training, thereby impeding their real-time applications. To circumvent this issue as well as improve the generalization capability of trained models, the use of stochastic training techniques has become widespread in deep learning community \cite{huang2016deep,lu2017beyond}. Unfortunately, design of the data-oriented network architecture for real-world learning problems is often more art than science, \textit{e.g.}, injection of the dropout layer into ResNet-like models \cite{he2016identity,zagoruyko2016wide}, tuning of the dropout probability and model hyperparameters \cite{huang2016deep,veit2016residual}, etc. As such, the latest studies have concentrated on the continuous-time dynamic of ResNets via (stochastic) modified equations \cite{li2017deep,lu2017beyond,weinan2017proposal}, which is highly desirable in order to explore the nature of deep ResNets and construct potentially more effective ones.

One essential objective of deep learning problems is to optimize their non-convex and highly rugged loss landscape \cite{dauphin2014identifying}, for which the stochastic gradient descent (SGD) method and its variants are commonly used. Although the optima found by SGD-based algorithms are of high quality measured by the training accuracy, it has been observed in practice \cite{chaudhari2016entropy,hoffer2017train,hu2017diffusion,keskar2016large} that sharp minima often lead to persistent degradations in test errors while flat minma tend to generalize well. Therefore, a number of regularization methods have emerged over the past few years to force the optimization toward flat plateau regions, among which the dropout approach \cite{srivastava2014dropout,wan2013regularization} has enjoyed widespread usage for its computing efficiency. Although the study in \cite{he2016identity} reported negative effects  of ResNet using dropout, more recent studies \cite{huang2016deep,zagoruyko2016wide} have shown that, by carefully inserting dropout layers into ResNet architectures, consistent gains on the generalization task can be achieved. However, the mechanism behind the demonstrate performance improvement has not been fully explored.

To better understand the dropout regularization effects as well as many other stochastic training methods, we propose in this work a ResNet-type architecture with dropout being inserted after the last convolutional layer of each building module. The method of stochastic modified equations is then applied to show its connection with a continuous-time stochastic differential equation. Since no dropout is applied at test time \cite{srivastava2014dropout}, the categorical scores are naturally described as the expectation of trained outputs, with the interpretation of evaluating an ensemble prediction over all the possible sub-networks \cite{srivastava2014dropout}. In turn, the It$\hat{\textnormal{o}}$'s formula is utilized to bridge the dropout training process with the optimal control of backward Kolmogorov's equations, where the training procedure of a plain ResNet can be interpreted as the optimization of systems constrained by transport equations. Consequently, from partial differential equations (PDEs) viewpoint \cite{evans2010partial}, the injected dropout layers act explicitly as a second-order artificial viscosity term to regularize the loss landscape, which cuts down the number of bad minimizers and lessens the energy barriers between pairs of local minima to aid optimization. This finding offers us a novel perspective on the regularization effects of a variety of stochastic training techniques such as Gaussian dropout \cite{srivastava2014dropout}, uniform dropout \cite{li2018understanding}, and shake-shake regularization \cite{gastaldi2017shake}. As a simple demonstration of the regularization properties,  the rugged loss landscape of a one-dimensional binary classification can be flattened out so that the network can escape from poor local minima and converge to flat minima. Then, experimental evaluations on a real-world image classification problem \cite{krizhevsky2009learning} are provided to further validate our theoretical findings. Though a higher training loss may be obtained, the generalization capability of the trained network is improved by using a suitable noise level, and the gap between training and inference procedures is significantly reduced. In other words, noise injection incurs a trade-off between model regularization and data fitting,  the determination of the optimal noise level associated with the depth and structure of ResNets is thus an important subject for further investigation. 

The rest of this paper is organized as follows. Section 2 is devoted to illustrating the notation used in this work, the plain ResNet architecture, and the method of modified equations. In Section 3, the ResNet with  inserted dropout layers is introduced, and a connection between the supervised learning task and the PDE-constrained optimization is established. More examples such as Gaussian dropout, uniform dropout, and shake-shake regularization are presented to show the wide applicability of this framework. A binary classification problem in dimension one is carried out in Section 4 to support our arguments, and experimental results on a real-world application  are reported as further validation. Conclusion remarks and future work are discussed in Section 6.


\section{Preliminaries}
This section is devoted to illustrating the notation used in this paper, the architecture of a plain ResNet associated with the image classification task and its connection with the modified equations.

\subsection{Notation}
The standard notation in \cite{evans2010partial,oksendal2003stochastic,higham2001algorithmic} is adopted in this paper to formulate the (stochastic) training processes of ResNet-like models as the PDE-constrained optimization problems. We shall denote by $\mathbb{Z}_n=\{0,\ldots,n-1\}$ the group of integers modulo $n\in\mathbb{N}$ and $\mathbb{Z}_n^m$ an $m\times 1$ vector whose elements belong to $\mathbb{Z}_n$. If $a=(a_1,\ldots,a_n)$ and $b=(b_1,\ldots,b_n)$ belong to $\mathbb{R}^{n\times1}$, $A=(a_{ij})$ and $B=(b_{ij})$ are $m\times n$ matrices, we define
\begin{equation*}
	a \cdot b = \sum_{i=1}^n a_i b_i \ \ \textnormal{and} \ \ A:B = \sum_{i=1}^m\sum_{j=1}^n a_{ij}b_{ij}.
\end{equation*}

The triplet $(\Omega,\mathcal{A},P)$ represents a complete probability space where $\Omega$ is a set of possible outcomes, $\mathcal{A}$ is the $\sigma$-algebra of events, and $P : \mathcal{F}\rightarrow[0,1]$ is the probability measure. Then, given a random variable $\omega$ distributed according to a certain probability distribution $\mathcal{N}$, the expectation and variance of a function $\ell(\omega)$ in variable $\omega\in\Omega$ are defined, respectively, by
\begin{equation*}
	\mathbb{E}_{\,\omega\sim \mathcal{N}} \! \left[ \ell(\omega) \right]=\int_\Omega \ell(\omega)\, dP(\omega)\ \ \textnormal{and}\ \ \mathbb{V}_{\omega\sim \mathcal{N}} \! \left[\ell(\omega)\right]=\mathbb{E}_{\,\omega\sim \mathcal{N}} \! \left[\big(\ell(\omega)-\mathbb{E}_{\, \omega\sim \mathcal{N}}[\ell(\omega)]\big)^2\right],
\end{equation*}
where the subscript $\omega\sim\mathcal{N}$ is retained throughout this paper to show under which distribution the expectation is being taken. In particular, if the random variable $\omega$ is independent and identically distributed over $\Omega$, the expected value of $\ell(\omega)$ is denoted by $\mathbb{E}_{\,\omega\in\Omega}[\ell(\omega)]$ for simplicity. Furthermore, $N(\mu,\nu^2)$ indicates a Gaussian random variable with mean $\mu$ and variance $\nu^2$, and the uniform distribution on interval $[\alpha,\beta]$ is written by $U(\alpha,\beta)$.

On the other hand, let $X_t$ evolve according to a stochastic differential equation \begin{equation*}
	dX_t = f(X_t,t)dt + g(X_t,t) dW_t, 
\end{equation*}
then for any $s\geq t$, the expectation of a xfunction $\ell(X_s)$ over all paths that originate from state $x$ at time $t$ is denoted by $\displaystyle \mathbb{E}\big(\ell(X_s)|X_t=x\big)$ \cite{oksendal2003stochastic}.

\subsection{Plain ResNet} \label{Section-Plain-ResNet}
In this work, we focus on the supervised learning problems, \textit{e.g.}, the image classification task over a given number of classes $m\in\mathbb{Z}^+$. ResNet is designed to infer the ground-truth label or, equivalently, the categorical distribution on one end for the input image given at the other.

Note that the adjacent pixels within natural images are strongly correlated (as is normally the case in early convolution layers where feature maps exhibit strong spatial correlation \cite{deng2009imagenet}), a convolutional layer with fixed or learned parameters is often utilized to map the raw image into its feature space \cite{Haber2017Learning}, followed by several stages each consisting of multiple building blocks \cite{he2016deep,he2016identity}. Since the input data usually has thousands of pixel values, \textit{e.g.}, an RGB image of size $32\times 32$ for the CIFAR dataset \cite{krizhevsky2009learning} and $256\times 256$ for the ImageNet dataset \cite{deng2009imagenet} after pre-processing, dimension reduction is carried out at transition stages of ResNet architecture for parameter-efficiency. After implementing the pooling operation, the network ends with a fully connected layer which reduces the output into a single vector of size $m\times1$ \cite{he2016deep}. In other words, ResNet maps the domain of complicated input images into a simpler but robust feature space so that the categorical scores can be easily obtained through a softmax classification layer \cite{he2016identity}. This allows us to measure the discrepancy between the ground-truth and predicted categorical distributions under a cross-entropy loss regime and, therefore, optimize the ResNet parameters via gradient-based algorithms.

For the ease of illustration, $n\in\mathbb{Z}^+$ denotes a universal constant whose value may change with the stage of feature flow, the dataset is assumed to contain the pre-processed image $y$ in a set of samples $\Omega\subset\mathbb{R}^d$ and its ground-truth label $h(y)\in\mathbb{Z}^m_2$, and the hyperparameters are manually set as those in benchmark models before the training process begins \cite{Haber2017Learning,li2017deep,lu2017beyond,weinan2017proposal}. More specifically, the first layer in a ResNet architecture is defined by the convolutional operation without bias terms:
\begin{equation}
	\mathcal{S}:\mathbb{R}^d \to \mathbb{R}^{n},\qquad y \mapsto Sy := X_0,
	\label{ResNet-Architecture-First-Layer}
\end{equation}
where $S\in\mathbb{R}^{d\times n}$ is a matrix to be learned, and the output feature map $X_0$ corresponds to the input of the subsequent building modules:
\begin{equation*}
	X_{k+1} = X_k + \mathcal{F}(X_k,w_k),\quad k=0,\ldots,K-1.
\end{equation*}
Here, $X_k$ indicates the input feature map of the $k$-th block, $\mathcal{F}$ typically a composition of linear and nonlinear functions, $w_k$ the module parameters to be trained, and $K\in\mathbb{Z}^+$ the total number of building blocks. 

\begin{remark}\label{Remark-BasicBlock}
For instance, the $BasicBlock$ in a full pre-activation ResNet (PreResNet) \cite{he2016identity} takes the form
\begin{equation*}
	\mathcal{F}(X_k,w_k) = w_k^{(2)}\cdot a\left(w_k^{(1)}\cdot a(X_k)\right) 
\end{equation*}
where $a=\textnormal{ReLU}\circ\textnormal{BN}$ is a composition of ReLU activation function and batch normalization (BN), $w_k:=\{w_k^{(1)},w_k^{(2)}\}$ represents the weight parameters in convolutional layers to be learned, and the parameters in BN layers are omitted for simplicity. One can also adopt the more economical residual block, \textit{e.g.}, the $Bottleneck$ module \cite{he2016deep}, and many others which are irrelevant to the purpose of this work.
\end{remark}

At the final layer, the input feature map is computed with a matrix multiplication followed by a bias offset, therefore leading to the single vector of categorical scores after softmax normalization, namely,
\begin{equation}
	\mathcal{T}:\mathbb{R}^n \to (0,1)^m,\qquad X_K \mapsto \mathcal{T}(X_K) = \sigma(AX_K+b),
	\label{Plain-ResNet-Discrete-Final-Layer}
\end{equation}
where $A\in\mathbb{R}^{n\times m}$ contains the classification weights, $b\in\mathbb{R}^m$ the bias vector, and $\sigma$ a softmax function (the pooling operation is ignored here, since the input feature map $X_T$ is already a vector). Consequently, the training procedure of a plain ResNet can be expressed as the minimization of a loss function
\begin{equation}
	\mathcal{J}\left(\mathcal{S},\mathcal{F},\mathcal{T}\right) = \mathbb{E}_{\, y\in\Omega} \Big[ \lVert \mathcal{T}(X_K) - h(y) \rVert \Big]
	\label{Plain-ResNet-Discrete-Loss}
\end{equation}
subject to the following constraints
\begin{equation}
	X_0 = \mathcal{S}y,\quad	X_{k+1} = X_k + \mathcal{F}(X_k,w_k),\quad k=0,\ldots,K-1.
	\label{Plain-ResNet-Discrete-DataFlow}
\end{equation}
Note that in \eqref{Plain-ResNet-Discrete-Loss}, $\lVert \cdot\rVert$ is a metric that measures the discrepancy between the ground-truth label and the model prediction for each training data, and the expectation is taken over all the samples in a given training database.

\subsection{Modified Equation}
The method of modified equations is widely used in numerical solution of PDEs, especially for finite difference schemes in the literature. Due to its ability to capture the effective behavior of numerical approximations, modified equation has become the primary technique for analyzing and improving finite difference algorithms \cite{li2017stochastic,warming1974modified}.

This methodology has been recently applied to gaining insight into the nature of ResNet-like architectures with or without noise  injection \cite{li2017deep,weinan2017proposal}, which also sheds light on the designing of more effective neural networks and explainable stochastic training strategies \cite{gastaldi2017shake,lu2017beyond}. To be specific, by conceptually introducing two positive parameters $\eta=\Delta t$, the feature flow associated with a plain ResNet \eqref{Plain-ResNet-Discrete-DataFlow} can be formulated as:
\begin{equation*}
	X_0 = \mathcal{S}y,\quad X_{k+1} = X_k + \frac{\mathcal{F}(X_k,w_k)}{\eta} \Delta t,\quad k=0,\ldots,K-1,
\end{equation*}
which is nothing but the forward Euler discretization of the following modified equations with time step $\Delta t=\eta>0$, that is,
\begin{equation}
	X_0 = \mathcal{S}y,\quad dX_t = f(X_t,t)d t, \quad t\in(0,1] \footnote{In contrast to the output of the first building block in a ResNet architeture \eqref{Plain-ResNet-Discrete-DataFlow}, $X_1$ (not relabeled) refers to the terminla value of modified equations throughout this work, where the subscript represents the terminal time $t=1$ for the ease of notation.},
	\label{Plain-ResNet-Continuous-DataFlow}
\end{equation}
where $f(X_t,t) := \mathcal{F}(X_t,w_t)/\eta$ and $\eta^{-1}=K$ is the number of building modules in a ResNet architecture \cite{chang2017multi,weinan2017proposal}. Note that the hyperparameters are manually defined prior to the commencement of the learning process, seeking the model parameters $\{w_k\}_{k=0}^{K-1}$ of a ResNet architecture \eqref{Plain-ResNet-Discrete-DataFlow} is equivalent to finding the function $f(x,t)$ of dynamic system \eqref{Plain-ResNet-Continuous-DataFlow} in the sense of modified equations.

This connection immediately allows the use of numerous existing numerical schemes regarding the discretization of ordinary differential equations \eqref{Plain-ResNet-Continuous-DataFlow}, and therefore constructs potentially more powerful architectures such as the linear multi-step ResNet \cite{lu2017beyond}. However, deeper network usually requires days or weeks for training, which makes it practically infeasible for online learning. To circumvent this issue as well as improve the generalization capability of the learned network, stochastic training strategies have enjoyed widespread usage in which the units or layers of a ResNet model are randomly dropped during training, \textit{e.g.}, ResNet with stochastic depth \cite{huang2016deep} which can be interpreted as the weak approximation of continuous-time stochastic modified equations \cite{lu2017beyond}. We refer the readers to \autoref{Section-Stochastic-Training} for more examples and a detailed analysis.


\section{Stochastic Training as PDE-constrained Optimization} \label{Section-Stochastic-Training}
In this section, we show that ResNet with injected noise can be regarded as numerical discretization of stochastic modified equations, which bridges a variety of stochastic training strategies with the optimal control of backward Kolmogorov's equations \cite{ito1974diffusion}. The underlying PDE interpretation enables us to study the stochastic regularization effects via the perspective of artificial viscosity \cite{evans2010partial}.

For the ease of illustration, we first propose a new way of utilizing Bernoulli dropout \cite{srivastava2014dropout} within the plain ResNet architecture such that a connection between the image classification task and the PDE-constrained optimization can be established. We should emphasize here that although the analysis is confined to Bernoulli dropout, this framework is applicable to various stochastic training methods including, but not limited to, Gaussian dropout, uniform dropout, and shake-shake regularization \cite{huang2016deep,lu2017beyond,gastaldi2017shake}, which helps understand and improve the stochastic regularization effects. More details can be found in section \ref{Section-NI-AV-Beyond} which support our aforesaid arguments.

\subsection{Dropout ResNet as Stochastic Modified Equations}
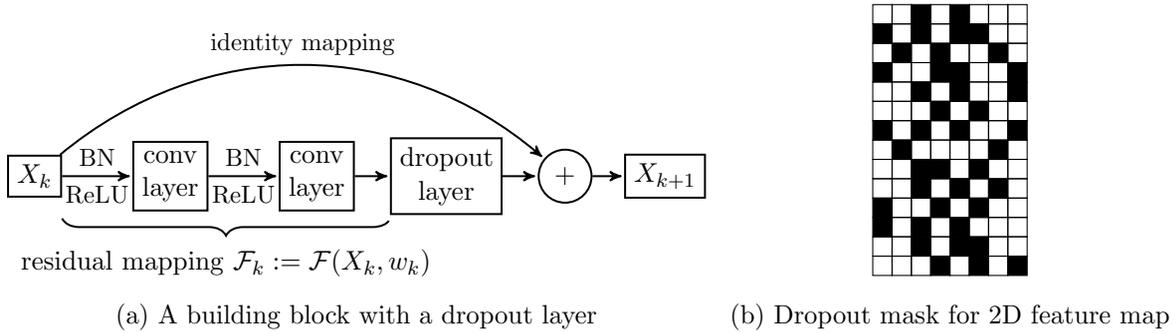
\begin{figure}[t!]
\centering
\begin{subfigure}[b]{0.61\linewidth}
\centering
\resizebox{\linewidth}{!}{
\begin{tikzpicture}[shorten >=1pt,auto,node distance=2.5cm,thick,main node/.style={rectangle,draw,font=\normalsize},decoration={brace,mirror,amplitude=7}]

\node[main node] (1) {$X_k$};
\node[main node] (2) [right =1cm of 1] {\parbox{0.8cm}{\centering conv \\ layer} };
\node[main node] (3) [right = 1cm of 2] {\parbox{0.8cm}{\centering conv \\ layer} };
\node[main node] (4) [right = 0.5cm of 3] {\parbox{1.32cm}{\centering dropout\\ layer} };
\node[draw,circle] (5) [right = 0.5cm of 4] {+};
\node[main node] (6) [right =0.45cm of 5]  {$X_{k+1}$};

\path[every node/.style={font=\sffamily\small},->,>=stealth']
    (1) edge node[above,midway] {\textnormal{BN}} node[below,midway] {\textnormal{ReLU}} (2)
    		edge [bend left=40] node[above] {\textnormal{identity mapping}} (5)
    (2) edge node[above,midway] {\textnormal{BN}} node[below,midway] {\textnormal{ReLU}} (3)
    (3) edge node {} (4)
    (4) edge node {} (5)
    (5) edge node {} (6);
    
\draw [decorate] ([yshift=-6mm]1.east) -- node[below=3mm]{residual mapping $\mathcal{F}_k:=\mathcal{F}(X_k,w_k)$} ([yshift=-6mm]4.west);

\end{tikzpicture}
}
\caption{A building block with a dropout layer}
\label{Fig-Block-PreResNet-Dropout}
\end{subfigure}
\hfill
\begin{subfigure}[b]{0.38\linewidth}
\centering
\begin{tikzpicture}[fill=black]
	\matrix[matrix of nodes, nodes={draw,minimum size=0.08cm}, nodes in empty cells,column sep=-\pgflinewidth,row sep=-\pgflinewidth](M){
               &          & |[fill]|     &             & |[fill]|     &             &          &          \\	
      |[fill]|  &          &   |[fill]|   &              &   |[fill]|   &   |[fill]|   &          &          \\
               & |[fill]|  &             &    |[fill]|   &             &             & |[fill]|  &          \\
      |[fill]|  &          &            &     |[fill]|   &   |[fill]|   &              &          & |[fill]| \\
               &          &   |[fill]|   &              &   |[fill]|   &              &          & |[fill]| \\
               &          &             &   |[fill]|    &             & |[fill]|     &          &          \\
      |[fill]|  &          & |[fill]|     &              & |[fill]|     &             &          & |[fill]| \\
               & |[fill]|  &              &              &              &            & |[fill]| &          \\
               &          & |[fill]|     & |[fill]|     &             & |[fill]|     &          &          \\
               &          & |[fill]|     &             & |[fill]|     &             & |[fill]|  &          \\ 
      |[fill]|  &          &             & |[fill]|     &              & |[fill]|     &          &          \\                             
      |[fill]|  &          & |[fill]|     &             & |[fill]|     &             &          &          \\
               &          & |[fill]|     &             & |[fill]|     & |[fill]|     &          &          \\
               &          &             & |[fill]|     &            & |[fill]|     &          & |[fill]|    \\                                             
               };       
\end{tikzpicture}
\caption{Dropout mask for 2D feature map}  
\label{Fig-PreResNet-M-Mask}
\end{subfigure}
\caption{The proposed building block with dropout layer being inserted, \textit{e.g.}, binary dropout layer using Bernoulli distribution.}
\label{Fig-ResNet-dropout}
\end{figure}

\def\layersep{1.5cm}
\begin{figure}[t!]
\centering
\begin{subfigure}[b]{0.33\textwidth}
\centering
\resizebox{\linewidth}{!}{
\begin{tikzpicture}[shorten >=1pt,>=latex,draw=black!100, node distance=\layersep,scale=1,cross/.style={path picture={ 
  \draw[black]
(path picture bounding box.south east) -- (path picture bounding box.north west) (path picture bounding box.south west) -- (path picture bounding box.north east);
}}]
    \tikzstyle{every pin edge}=[<-,shorten <=1pt,]
    \tikzstyle{neuron}=[circle,fill=black!25,minimum size=12pt,inner sep=0pt]
    \tikzstyle{input neuron}=[neuron, fill=white!100,draw=black];
    \tikzstyle{output neuron}=[neuron, fill=white!100,draw=black];
    \tikzstyle{hidden neuron}=[neuron, fill=white!100,draw=black];
    \tikzstyle{annot} = [text width=10em, text centered]
    \tikzstyle{dropout}=[circle,cross,fill=black!25,minimum size=12pt,inner sep=0pt]
    
\foreach \name / \y in {1,2,3,4,5}
    \node[input neuron, pin=left:] (I-\name) at (0,-\y) {};

\foreach \name / \y in {1,...,5}
    \path node[hidden neuron] (H-\name) at (\layersep,-\y cm) {};

\node[output neuron,pin={[pin edge={->}]right:}, right of=H-1,label={$\mathcal{F}_k$}] (O1) {};
\node[hidden neuron,left of=O1,label={}] (H-1) {};
\node[input neuron,left of=H-1,label={$X_k$}] (I-1) {};

\node[output neuron, right of=H-1] {};
\node[dropout, right of=H-1] (O1) {};
\node[output neuron, right of=H-2] {};
\node[dropout, right of=H-2] (O2) {};
\node[output neuron,pin={[pin edge={->}]right:}, right of=H-3] (O3) {};
\node[output neuron, right of=H-4] {};
\node[dropout, right of=H-4] (O4) {};
\node[output neuron,pin={[pin edge={->}]right:}, right of=H-5] (O5) {};

\foreach \source in {1,2}
        \path[->] (I-\source) edge (H-1);
\foreach \source in {1,2,3}
        \path[->] (I-\source) edge (H-2);  
\foreach \source in {2,3,4}
        \path[->] (I-\source) edge (H-3);               
\foreach \source in {3,4,5}
        \path[->] (I-\source) edge (H-4);
\foreach \source in {4,5}
        \path[->] (I-\source) edge (H-5);        
        
\foreach \source in {4,5}
    \path[->] (H-\source) edge (O5);
\foreach \source in {2,3,4}
    \path[->] (H-\source) edge (O3);

\node[output neuron, left of=H-5] {};
\node[output neuron, left of=H-2] {};
\node[output neuron, left of=H-3] {};
\end{tikzpicture}
}
\caption{Dropout residual mapping}   
\label{Fig-Bernoulli-Dropout-Network}
\end{subfigure}
\qquad
\begin{subfigure}[b]{0.54\linewidth}
\centering
\resizebox{\linewidth}{!}{
\begin{tikzpicture}[shorten >=1pt,auto,node distance=2.5cm,thick,main node/.style={rectangle,draw,font=\normalsize},decoration={brace,mirror,amplitude=7}]

\node[draw=none] (11) {$X_k$};
\node[main node,gray] (12) [right =0.4cm of 11] {\parbox{0.9cm}{\centering $\mathcal{F}_k$} };
\node[draw,circle,gray] (13) [right = 0.4cm of 12] {};
\node[main node,gray] (14) [right = 0.4cm of 13] {\parbox{0.9cm}{\centering $\mathcal{F}_{k+1}$} };
\node[draw,circle] (15) [right = 0.4cm of 14] {};
\node[main node] (16) [right = 0.4cm of 15] {\parbox{0.9cm}{\centering $\mathcal{F}_{k+2}$} };
\node[draw,circle] (17) [right = 0.4cm of 16] {};
\node[draw=none] (18) [right = 0.4cm of 17] {$X_{k+3}$};

\path[every node/.style={font=\sffamily\small},->,>=stealth',gray]
    (11) edge node[above,midway] {} node[below,midway] {} (12)
    (12) edge node[above,midway] {} node[below,midway] {} (13)
    (13) edge node[above,midway] {} node[below,midway] {} (14)
    (14) edge node[above,midway] {} node[below,midway] {} (15);
\path[every node/.style={font=\sffamily\small},->,>=stealth']   
    (15) edge node[above,midway] {} node[below,midway] {} (16)
    (16) edge node[above,midway] {} node[below,midway] {} (17)
    (17) edge node[above,midway] {} node[below,midway] {} (18);

\path[draw,->,>=stealth',gray] 
    (11.north) -- ++(0,0.4cm) -- ++(2.57cm,0) -- (13.north);
    
\node[main node] (22) [above =0.8cm of 12] {\parbox{0.9cm}{\centering $\mathcal{F}_k$} };
\node[draw,circle] (23) [right = 0.4cm of 22] {};

\path[draw,->,>=stealth'] 
	(22) edge node[above,midway] {} node[below,midway] {} (23);
\path[draw,->,>=stealth'] 
    (11.north) -- ++(0,1.1cm)  -- (22.west);
\path[draw,->,>=stealth']    
    (23.east)  -- ++(2.16cm,0) -- (15.north);

\path[draw,->,>=stealth'] 
    (11.north) -- ++(0,1.8cm) -- ++(2.57cm,0) -- (23.north);
    
\node[main node,gray] (32) [above =0.8cm of 22] {\parbox{0.9cm}{\centering $\mathcal{F}_k$} };
\node[draw,circle,gray] (33) [right = 0.4cm of 32] {};
\node[main node,gray] (34) [right =0.4cm of 33] {\parbox{0.9cm}{\centering $\mathcal{F}_{k+1}$} };
\node[draw,circle] (35) [right = 0.4cm of 34] {};

\path[draw,->,>=stealth',gray] 
	(32) edge node[above,midway] {} node[below,midway] {} (33)
	(33) edge node[above,midway] {} node[below,midway] {} (34)
	(34) edge node[above,midway] {} node[below,midway] {} (35);
\path[draw,->,>=stealth',gray] 
    (11.north) -- ++(0,2.51cm)  -- (32.west);
\path[draw,->,>=stealth'] 
    (35.east) -- ++(2.15cm,0)  -- (17.north);
    
\path[draw,->,>=stealth',gray] 
    (11.north) -- ++(0,3.2cm) -- ++(2.57cm,0) -- (33.north);
    
\node[main node] (42) [above =0.8cm of 32] {\parbox{0.9cm}{\centering $\mathcal{F}_k$} };
\node[draw,circle] (43) [right = 0.4cm of 42] {};

\path[draw,->,>=stealth'] 
	(42) edge node[above,midway] {} node[below,midway] {} (43)
	(43.east)  -- ++(2.16cm,0) -- (35.north);
\path[draw,->,>=stealth'] 	
	(11.north) -- ++(0,3.92cm)  -- (42.west);
    
\path[draw,->,>=stealth'] 
    (11.north) -- ++(0,4.58cm) -- ++(2.57cm,0) -> (43.north);
    
\end{tikzpicture}
}
\caption{An unraveled view of a 3-block PreResNet}
\label{SubFig-ResNet-Unraveled-View}
\end{subfigure}

\caption{(a) The residual mapping using dropout during training. (b) The unraveled view of a 3-block ResNet with residual block $\mathcal{F}_{k+1}$ dropped \cite{veit2016residual}, \textit{i.e.}, $\gamma_{k}=\gamma_{k+2}=I=I-\gamma_{k+1}$.}
\label{Fig-Dropout-Network}
\end{figure}
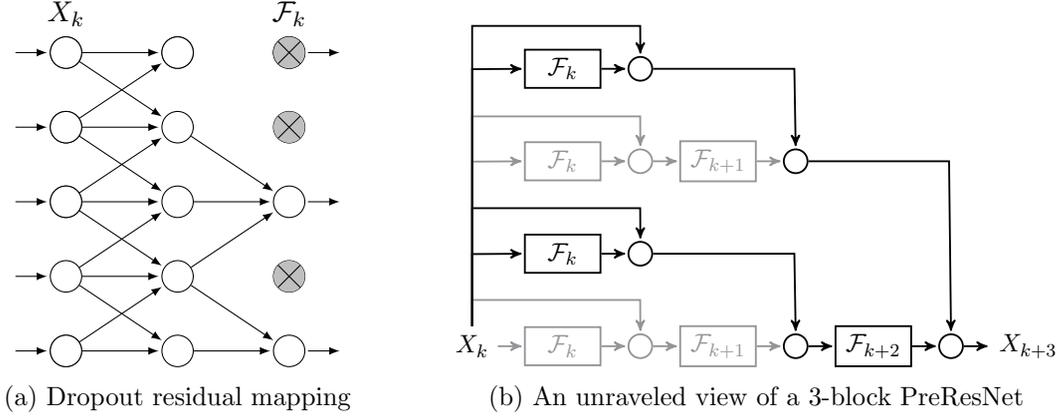

Without loss of generality, we consider here the modified PreResNet in which a single dropout layer is assigned after the last convolutional layer in each building block as depicted \autoref{Fig-Block-PreResNet-Dropout}. Then the feature flow of the proposed network during training can be described as:
\begin{equation*}
	X_0 = \mathcal{S}y,\quad X_{k+1} = X_k + \mathcal{F}(X_k,w_k) \odot \gamma_k,\quad k=0,\ldots,K-1,
\end{equation*}
where $\odot$ is the element-wise product, and $\gamma_k\in\mathbb{Z}_2^n$ represents a binary mask vector with each element drawn independently from a Bernoulli distribution $\textnormal{Bern}(p)$ with survival probability $p\in(0,1]$ during iteration. That is, for every component of $\gamma_k$ (not relabeled), we have 
\begin{equation*}
	P\left(\gamma_k=1\right) = p = 1 - P\left(\gamma_k=0\right).
\end{equation*}
Put differently, the binary vector $\gamma_k$ determines which parts of the feature map are taken into account or dropped in the subsequent blocks. Moreover, by expanding the ResNet in an unraveled view as depicted in \autoref{SubFig-ResNet-Unraveled-View}, \cite{veit2016residual} states that one do not need a specific policy, \textit{e.g.}, stochastic depth \cite{huang2016deep}, to determine the optimal survival probability associated with the depth of feature flow. As such, the survival probability is set uniformly for all building blocks to get a single model hyperparameter $0<p\leq 1$ throughout this work.

At test time, the trained units need to be scaled down by a factor of $p>0$ in order to attain the ensemble prediction of exponentially many sub-networks  \cite{srivastava2014dropout,veit2016residual}. To achieve the same effect, one can scale up the retained activations by $1/p$ at training time, \textit{i.e.},
\begin{equation}
	X_{k+1} = X_k + \mathcal{F}(X_k,w_k) \odot \frac{\gamma_k}{p},
	\label{Dropout-ResNet-Discrete-DataFlow}
\end{equation}
and not modify the parameters in inference \cite{srivastava2014dropout}. By incorporating the parameters $\eta=\Delta t>0$ and letting $I$ be an all-ones vector that has the same size as $\gamma_k$, formula \eqref{Dropout-ResNet-Discrete-DataFlow} can be rewritten as:
\begin{equation}
\begin{split}
	X_{k+1} & = X_k + \mathcal{F}(X_k,w_k) + \mathcal{F}(X_k,w_k) \odot \left( \frac{\gamma_k}{p} - I \right) \\
	& = X_k + \frac{\mathcal{F}(X_k,w_k)}{\eta} \Delta t + \frac{\sqrt{\eta (1-p)}}{\sqrt{p}} \frac{\mathcal{F}(X_k,w_k)}{\eta} \odot \frac{(\gamma_k-p \cdot I)\sqrt{\Delta t}}{\sqrt{p(1-p)}} \\
	& = X_k + \frac{\mathcal{F}(X_k,w_k)}{\eta} \Delta t + \frac{\sqrt{\eta (1-p)}}{\sqrt{p}} \frac{\mathcal{G}(X_k,w_k)}{\eta} \frac{(\gamma_k-p \cdot I)\sqrt{\Delta t}}{\sqrt{p(1-p)}} 
\end{split}	
\label{Dropout-ResNet-Discrete-DataFlow-Reformulated}
\end{equation}
where $\mathcal{G}(X_k,w_k)$ is a diagonal matrix with diagonal entries defined by $\mathcal{F}(X_k,w_k)$, and $p\cdot I$ is the product of a scalar $p$ and a vector $I$.

Note that the expectation and variance of the random vector in \eqref{Dropout-ResNet-Discrete-DataFlow-Reformulated} satisfy
\begin{equation}
	\mathbb{E}_{\, \gamma_k\sim\textnormal{Bern}(p)} \! \left[ \frac{(\gamma_k-p \cdot I)\sqrt{\Delta t}}{\sqrt{p(1-p)}} \right] = 0\cdot I\ \ \textnormal{and}\ \ \mathbb{V}_{\gamma_k\sim\textnormal{Bern}(p)} \! \left[ \frac{(\gamma_k-p \cdot I)\sqrt{\Delta t}}{\sqrt{p(1-p)}} \right] = \Delta t\cdot I,
	\label{Dropout-ResNet-Discrete-DataFlow-Statistics}
\end{equation}
which implies that \eqref{Dropout-ResNet-Discrete-DataFlow-Reformulated} is a weak approximation of the stochastic modified equations \cite{Kloeden2010The,lu2017beyond}
\begin{equation}
	X_0=\mathcal{S}y, \quad	dX_t = f(X_t,t)dt + \varepsilon g(X_t,t) dW_t, \quad t\in(0,1],
	\label{Dropout-ResNet-Continuous-DataFlow}
\end{equation}
where $\varepsilon^2 := \eta \nu^2$, $\nu^2 := (1-p)/p = \mathbb{V}_{\gamma_k\sim\textnormal{Ben}(p)}[\gamma_k/p-I]$, $g(X_t,t):=\mathcal{G}(X_t,w_t)/\eta$, and $W_t$ indicates the multidimensional Brownian motion \cite{oksendal2003stochastic}.

\subsection{Dropout Training as Optimal Control of PDEs}
Recall that at the final layer, the input feature map or, equivalently, the terminal value $X_1$ in \eqref{Dropout-ResNet-Continuous-DataFlow} is evaluated by the mapping $\mathcal{T}(\cdot)$ in classification problems to get the categorical scores $\mathcal{T}(X_1)$. Moreover, there is no dropout applied at test time \cite{srivastava2014dropout}, namely, $p=1$ or, equivalently, $\varepsilon^2=0$ in \eqref{Dropout-ResNet-Continuous-DataFlow}, with the interpretation of evaluating an ensemble prediction over all the possible sub-networks with shared parameters \cite{srivastava2014dropout}. As such, by defining the conditional expectation
\begin{equation}
	u(x,t) = \mathbb{E} \Big( \mathcal{T}(X_1)\,|\,X_t=x  \Big), \quad t\in[0,1],
	\label{Label-Prediction-Conditional-Expectation}
\end{equation}
where the expectation in \eqref{Label-Prediction-Conditional-Expectation} is taken over all sample paths of $X_1$ that originate from state $x$ at time $t$ \cite{oksendal2003stochastic}, it follows immediately from \eqref{Dropout-ResNet-Continuous-DataFlow}, \eqref{Label-Prediction-Conditional-Expectation}, and \eqref{Plain-ResNet-Discrete-Final-Layer} that $u(\mathcal{S}y,0)$ serves as the predicted categorical distribution for the input image $y\in\Omega$.

On the other hand, by the multidimensional It$\hat{\textnormal{o}}$'s formula \cite{ito1974diffusion}, function \eqref{Label-Prediction-Conditional-Expectation} is known to solve the backward Kolmogorov's equation 
\begingroup
\renewcommand*{\arraystretch}{1.5}
\begin{equation}
\left\{
\begin{array}{ll}
\displaystyle u_t + f\cdot\nabla u + \frac{\varepsilon^2}{2} gg^T:\nabla^2 u = 0\ &\ \textnormal{in}\ \ \Omega\times[0,1),\\
u(x,1) = \mathcal{T}(x)\ &\ \textnormal{in}\ \ \Omega.
\end{array}\right.
\label{Backward-Kolmogorov-Equation}
\end{equation}
\endgroup
Consequently, for a classification problem with ground-truth label function $h(y)$, by introducing the loss function
\begin{equation}
	\mathcal{J}_\varepsilon \left( \mathcal{S},f,\mathcal{T} \right) = \mathbb{E}_{\, y\in\Omega} \Big[ \|u(\mathcal{S}y,0)-h(y)\| \Big],
	\label{Loss-Function-Dropout-ResNet}
\end{equation}
the training process for the proposed ResNet with injected dropout layers is defined as:
\begin{equation}
	\textnormal{seek}\ \mathcal{S},f,\mathcal{T}\ \textnormal{such that}\ \eqref{Loss-Function-Dropout-ResNet}\ \textnormal{is minimized subject to constraint}\ \eqref{Backward-Kolmogorov-Equation}.
	\label{Opt-Ctl-PDE-Dropout-ResNet}
\end{equation}
In particular, the function $g$ in \eqref{Dropout-ResNet-Continuous-DataFlow} is diagonal and so is $g^T$, which implies that the high-order term $gg^T :\nabla^2 u$ in \eqref{Backward-Kolmogorov-Equation} is an anisotropic Laplacian, \textit{i.e.}, there is no mixed derivative terms.

To sum up, the method of stochastic modified equations enables us to establish a connection between training the ResNet with dropout layers being inserted and the optimal control of backward Kolmogorov's equations, which is highly desirable in order to understand and improve the dropout regularization effects. More discussions are carried out in the following sections to further illustrate the significance of this PDE interpretation.

\subsection{Dropout Regularization as Artificial Viscosity}
Based on the analysis established in the previous section, we are now able to reformulate the training procedure of a plain ResNet as the optimal control of transport equations. That is, by setting $p=1$ in \eqref{Dropout-ResNet-Continuous-DataFlow} (\textit{i.e.}, there is no dropout layers inserted in the ResNet architecture), the parabolic constraint \eqref{Backward-Kolmogorov-Equation} reduces to a hyperbolic system
\begingroup
\renewcommand*{\arraystretch}{1.5}
\begin{equation}
\left\{
\begin{array}{ll}
\displaystyle v_t + f\cdot\nabla v = 0\ &\ \textnormal{in}\ \ \Omega\times[0,1),\\
v(x,1) = \mathcal{T}(x)\ &\ \textnormal{in}\ \ \Omega,
\end{array}\right.
\label{Backward-Transport-Equation}
\end{equation}
\endgroup
since $\varepsilon = \eta \mu^2$ and $\mu^2 = (1-p)/p = 0$. Therefore, the control objective of a plain ResNet can be expressed as the minimization of a loss function
\begin{equation}
	\mathcal{J}_0(\mathcal{S},f,\mathcal{T}) = \mathbb{E}_{\, y\in\Omega} \Big[ \|v(\mathcal{S}y,0)-h(y)\| \Big]
	\label{Loss-Function-Plain-ResNet}
\end{equation}
subject to the transport equations \eqref{Backward-Transport-Equation} or, equivalently, 

\begin{equation}
	\textnormal{seek}\ \mathcal{S},f,\mathcal{T}\ \textnormal{such that}\ \eqref{Loss-Function-Plain-ResNet}\ \textnormal{is minimized subject to constraint}\ \eqref{Backward-Transport-Equation}.
	\label{Opt-Ctl-PDE-Plain-ResNet}
\end{equation}

\begin{remark}
Such a model problem can also be derived through the method of characteristics, in which the modified equations \eqref{Plain-ResNet-Continuous-DataFlow} associated with the feature flow of plain ResNets serves as the characteristic line of \eqref{Backward-Transport-Equation}. We refer the readers to reference \cite{li2017deep,evans2010partial} for more details. Note also that the functional given by \eqref{Loss-Function-Plain-ResNet} can be formally seen as the zero viscosity limit ($\epsilon=0$) of the one given in \eqref{Loss-Function-Dropout-ResNet}.
\end{remark}

As mentioned before, an anisotropic Laplacian appears in constraint \eqref{Backward-Kolmogorov-Equation} as compared to the plain ResNet \eqref{Backward-Transport-Equation}. Mathematically, such a term becomes relevant only when $\nabla^2 u$ or $g$ is large, that is, in the region where $\nabla u$ changes rapidly or $f$ is sufficiently large. This makes the PDE \eqref{Backward-Kolmogorov-Equation} act somewhat like the heat equations \cite{evans2010partial}, in spite of the nonlinearity, such as the ReLU activation function, in the building module $\mathcal{F}(X_k,w_k)$ and hence the velocity field $f(x,t)$. Physically, we can regard this term as imposing an artificial viscosity to the hyperbolic system \eqref{Backward-Transport-Equation}, where $u(x,t)$ indicates the viscous approximation of $v(x,t)$ \cite{evans2010partial}. 

Due to the softmax normalization in the terminal condition $\mathcal{T}(\cdot)$, another key observation is that the label prediction for both the plain ResNet \eqref{Opt-Ctl-PDE-Plain-ResNet} and dropout ResNet \eqref{Opt-Ctl-PDE-Dropout-ResNet} may have sharp contrast at the terminal time $t=1$. For instance, let $m=2$ and the model parameters $A=(a_0,a_1)^T$, $b=(b_0,b_1)^T$ in \eqref{Plain-ResNet-Discrete-Final-Layer}, where $a_i\in\mathbb{R}^{n\times 1}$ and $b_i\in\mathbb{R}$ for $0\leq i\leq 1$, then the final layer associated with a binary classification task takes on the form
\begin{equation}
	\mathcal{T}(X_K) = 
	\begin{pmatrix}
		\displaystyle \frac{1}{1+e^{-(a_0^T-a_1^T)X_K-(b_0-b_1)}} \\
		\displaystyle 1 - \frac{1}{1+e^{-(a_0^T-a_1^T)X_K-(b_0-b_1)}} \\
	\end{pmatrix}.
	\label{Binary-Classification-Prob-Terminal-Condition}
\end{equation}
Clearly, both the components of terminal condition $\mathcal{T}(\cdot)$ are $S$-shaped sigmoid functions that may have sharp contrast. As a result, the loss landscape for plain ResNet \eqref{Opt-Ctl-PDE-Plain-ResNet} can be highly non-convex and rugged since $v(x,t)$ remains constant along the characteristic line \eqref{Plain-ResNet-Continuous-DataFlow}. On the contrary, the viscous solution $u(x,t)$ becomes flattened out for $t<1$, therefore leading to a regularized loss landscape \eqref{Loss-Function-Dropout-ResNet} which brings down the poor local minima while almost retains the flat minima to close the generalization gap (see \autoref{Section-Loss-Landscape} for more details).

\subsection{Noise Injection, Artificial Viscosity, and Beyond} 
\label{Section-NI-AV-Beyond}
Before a comprehensive characterization of the stochastic regularization effects, several noise injection techniques are studied in this section to illustrate how the proposed theory helps gaining insight into the nature of stochastic training of ResNet-like models.

For instance, the dropout ResNets \eqref{Dropout-ResNet-Discrete-DataFlow-Reformulated} can be generalized by multiplying the activations with a Gaussian random variable $\gamma_k\sim N(1,\nu^2)$ (not relabeled) \cite{srivastava2014dropout}. Then, the flow of feature map during training can be expressed as:
\begin{equation}
\begin{split}
	X_{k+1} & = X_k + \mathcal{F}(X_k,w_k) \odot \gamma_k \\
	& = X_k + \frac{\mathcal{F}(X_k,w_k)}{\eta} \Delta t + \nu\sqrt{\eta} \frac{\mathcal{F}(X_k,w_k)}{\eta} \odot \frac{(\gamma_k-I) \sqrt{\Delta t}}{\nu} \\
	& = X_k + \frac{\mathcal{F}(X_k,w_k)}{\eta} \Delta t + \nu\sqrt{\eta} \frac{\mathcal{G}(X_k,w_k)}{\eta} \frac{(\gamma_k-I) \sqrt{\Delta t}}{\nu},
\end{split}	
\label{Gaussian-Dropout}
\end{equation}
which is the Euler-Maruyama discretization \cite{higham2001algorithmic} of stochastic modified equations \eqref{Dropout-ResNet-Continuous-DataFlow} with time step $\Delta t=\eta$ and parameter $\varepsilon=\nu\sqrt{\eta}$ due to the fact that
\begin{equation*}
	\mathbb{E}_{\,\gamma_k\sim N(1,\nu^2)} \! \left[ \frac{(\gamma_k-I) \sqrt{\Delta t}}{\nu} \right] = 0\cdot I\ \ \textnormal{and}\ \ \mathbb{V}_{\gamma_k\sim N(1,\nu^2)} \! \left[ \frac{(\gamma_k-I) \sqrt{\Delta t}}{\nu} \right] = \Delta t\cdot I.
\end{equation*}
Apparently the parameter $\nu$ is set equal to zero or, equivalently, $\varepsilon=0$ in \eqref{Dropout-ResNet-Continuous-DataFlow} at test time to attain the ensemble prediction \eqref{Label-Prediction-Conditional-Expectation}, which coincides with the practical implementation where no Gaussian dropout is used in inference.

Similar in spirit is the uniform dropout \cite{li2018understanding}, where a random variable $\gamma_k \sim U(-\beta,\beta)$ (not relabeled) is adopted in the feature flow during training:
\begin{equation}
\begin{split}
	X_{k+1} & = X_k + \mathcal{F}(X_k,w_k) \odot (I+\gamma_k) \\
	& = X_k + \frac{\mathcal{F}(X_k,w_k)}{\eta} \Delta t + \frac{\beta\sqrt{\eta}}{\sqrt{3}} \frac{\mathcal{G}(X_k,w_k)}{\eta} \frac{\sqrt{3\Delta t}}{\beta} \gamma_k ,
\end{split}	
\label{Uniform-Dropout}
\end{equation}
which is the weak approximation \cite{Kloeden2010The} of stochastic modified equations \eqref{Dropout-ResNet-Continuous-DataFlow} with time step $\Delta t$ $=\eta$ and parameter $\varepsilon = \frac{\beta\sqrt{\eta}}{\sqrt{3}}$ since
\begin{equation*}
	\mathbb{E}_{\,\gamma_k\sim U(-\beta,\beta)} \! \left[ \frac{\sqrt{3\Delta t}}{\beta} \gamma_k \right] = 0\cdot I\ \ \textnormal{and}\ \ \mathbb{V}_{\gamma_k\sim U(-\beta,\beta)} \! \left[ \frac{\sqrt{3\Delta t}}{\beta} \gamma_k \right] = \Delta t\cdot I.
\end{equation*}
On the other hand, the random variable $\gamma_k$ is set to be zero in test mode \cite{li2018understanding} to improve the accuracy, which is straightforward and easy to illustrate by our theory.

In addition to the aforementioned architectures, a 3-branch ResNet with noise injection is introduced in \cite{gastaldi2017shake}, which achieves the best single shot published results on the CIFAR dataset. More specifically, let $\mathcal{F}_i(X):=\mathcal{F}(X,w^{{(i)}})$ be the residual mapping characterized by parameter $w^{(i)}$ for $1\leq i\leq 2$ and $\gamma_k\sim U(0,1)$ denote a scalar random variable (not relabeled), then the training procedure using shake-shake regularization \cite{gastaldi2017shake} takes  on the form
\begin{equation}
\begin{split}
	X_{k+1} & = X_k + \mathcal{F}_1(X_k) \gamma_k + \mathcal{F}_2(X_k) (1-\gamma_k) \\
	& = X_k + \frac{\mathcal{F}_1(X_k)+\mathcal{F}_2(X_k)}{2} + \frac{\mathcal{F}_1(X_k)-\mathcal{F}_2(X_k)}{2} \left(2\gamma_k-1\right) \\
	& = X_k +\frac{\mathcal{F}_1(X_k)+\mathcal{F}_2(X_k)}{2\eta} \Delta t + \frac{\sqrt{\eta}}{\sqrt{3}} \frac{\mathcal{F}_1(X_k)-\mathcal{F}_2(X_k)}{2\eta} \left(2 \gamma_k-1\right)\sqrt{3\Delta t},
\end{split}
\label{Shake-Shake-Regularization}
\end{equation}
which is a weak approximation of \eqref{Dropout-ResNet-Continuous-DataFlow} with time step $\eta=\Delta t$, $g(X_t,t)=\textnormal{diag}\left(\frac{\mathcal{F}_1(X_t)-\mathcal{F}_2(X_t)}{2\eta}\right)$, $f(X_t,t)=\frac{\mathcal{F}_1(X_t)+\mathcal{F}_2(X_t)}{2\eta}$, and parameter $\varepsilon=\frac{\sqrt{\eta}}{\sqrt{3}}$ due to 
\begin{equation*}
	\mathbb{E}_{\, \gamma_k\sim U(0,1)} \! \left[ \left(2\gamma_k-1\right)\sqrt{3\Delta t} \right] = 0\ \ \textnormal{and}\ \ \mathbb{V}_{\gamma_k\sim U(0,1)} \! \left[ \left(2 \gamma_k-1\right)\sqrt{3\Delta t} \right] = \Delta t.
\end{equation*}
At test time, the parameter $\gamma_k$ is set to be $0.5$ \cite{gastaldi2017shake} which validates our analysis.

To conclude, the method of stochastic modified equations shows that the stochastic training strategies \eqref{Gaussian-Dropout}, \eqref{Uniform-Dropout}, and \eqref{Shake-Shake-Regularization} can be regarded as inserting multiplicative noise terms into the feature flow of their plain ResNet counterparts. The proposed PDE interpretation is thus applicable, where the injected noise acts explicitly as a second-order artificial viscosity term during the training process \eqref{Backward-Kolmogorov-Equation} to aid optimization. Although the Gaussian dropout has the highest entropy among all distributions of equal variance \cite{srivastava2014dropout}, the Bernoulli dropout seems preferable in the deep learning community due to its parameter-efficiency, \textit{i.e.}, a smaller number of model parameters are computed per-iteration (see \autoref{Fig-Bernoulli-Dropout-Network}) when compared with the Gaussian and uniform dropout networks.

\section{A Landscape Perspective on Stochastic Training}
\label{Section-Loss-Landscape}
To further characterize the regularization effects of stochastic training algorithms, the loss landscape perspective is used in this section to illustrate the potential boost in test accuracy by revealing the difference between minimizers found by \eqref{Opt-Ctl-PDE-Plain-ResNet} and \eqref{Opt-Ctl-PDE-Dropout-ResNet}. That is, by injecting a suitable amount of noise into the plain ResNet architecture \eqref{Plain-ResNet-Discrete-DataFlow}, the highly non-convex and rugged loss landscape \eqref{Plain-ResNet-Discrete-Loss} can be flattened out which offers the capability of escaping from poor local optima. We theoretically justify this finding through a binary classification problem in one dimension, while extensive experimental results for a real-world application support our arguments in the general high-dimensional case.

More specifically, recent evidences from practical applications \cite{keskar2016large} showed that although the multiple local minima have nearly equivalent training accuracy, the flat minimizers tend to generalize well while the sharp minimizers always lead to a degradation in generalization performance. As such, closing the gap in accuracy between training and testing procedures has received significant attention in recent years, which is the focus of this section.

\subsection{Binary Classification Task in Dimension One}
\begin{figure*}[t!]
    \centering
    \begin{subfigure}[t]{0.45\textwidth}
        \centering
        \includegraphics[width=0.9\textwidth]{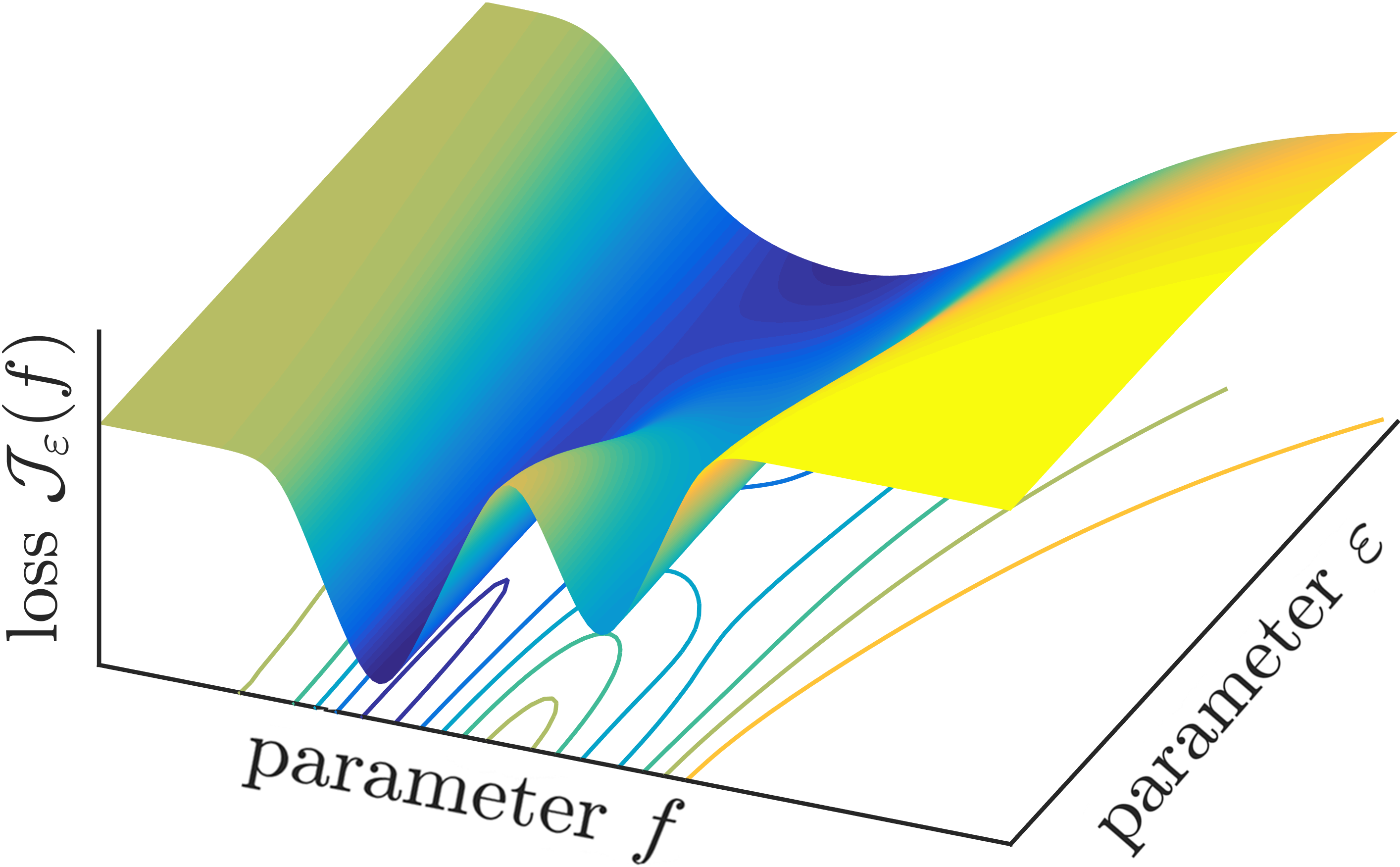}
        \caption{$\ell_2$-loss landscape w.r.t. $\varepsilon$ and $f$}
    \end{subfigure}%
    ~ 
    \begin{subfigure}[t]{0.45\textwidth}
        \centering
        \includegraphics[width=0.9\textwidth]{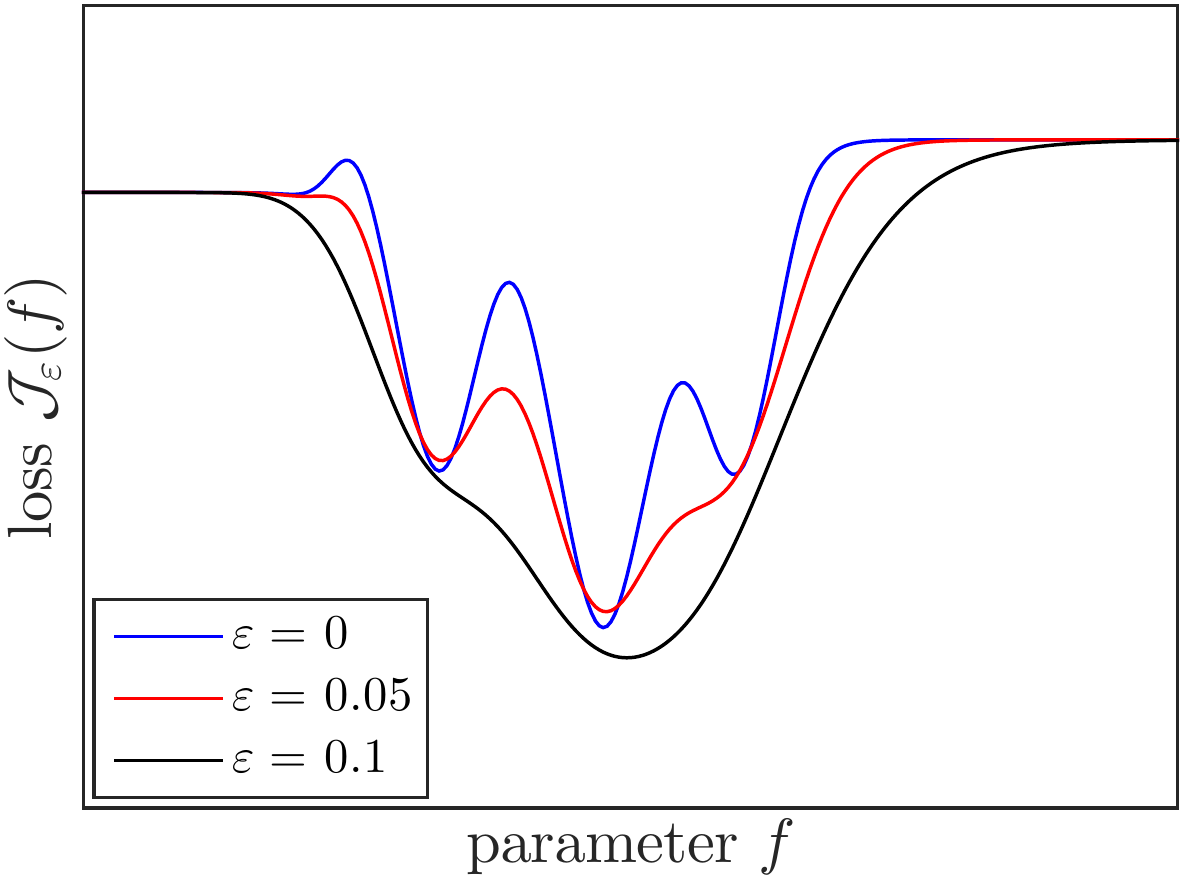}
        \caption{Slices of $\ell_2$-loss landscape}
    \end{subfigure}
    \caption{Noise injection can smooth the $\ell_2$-loss landscape to aid optimization, \textit{e.g.}, lessening of energy barrier between pairs of minima and enhancing the convexity of rugged landscape.}
    \label{Fig-L2-loss-const}
\end{figure*}

\begin{figure*}[t!]
    \centering
    \begin{subfigure}[t]{0.45\textwidth}
        \centering
        \includegraphics[width=0.9\textwidth]{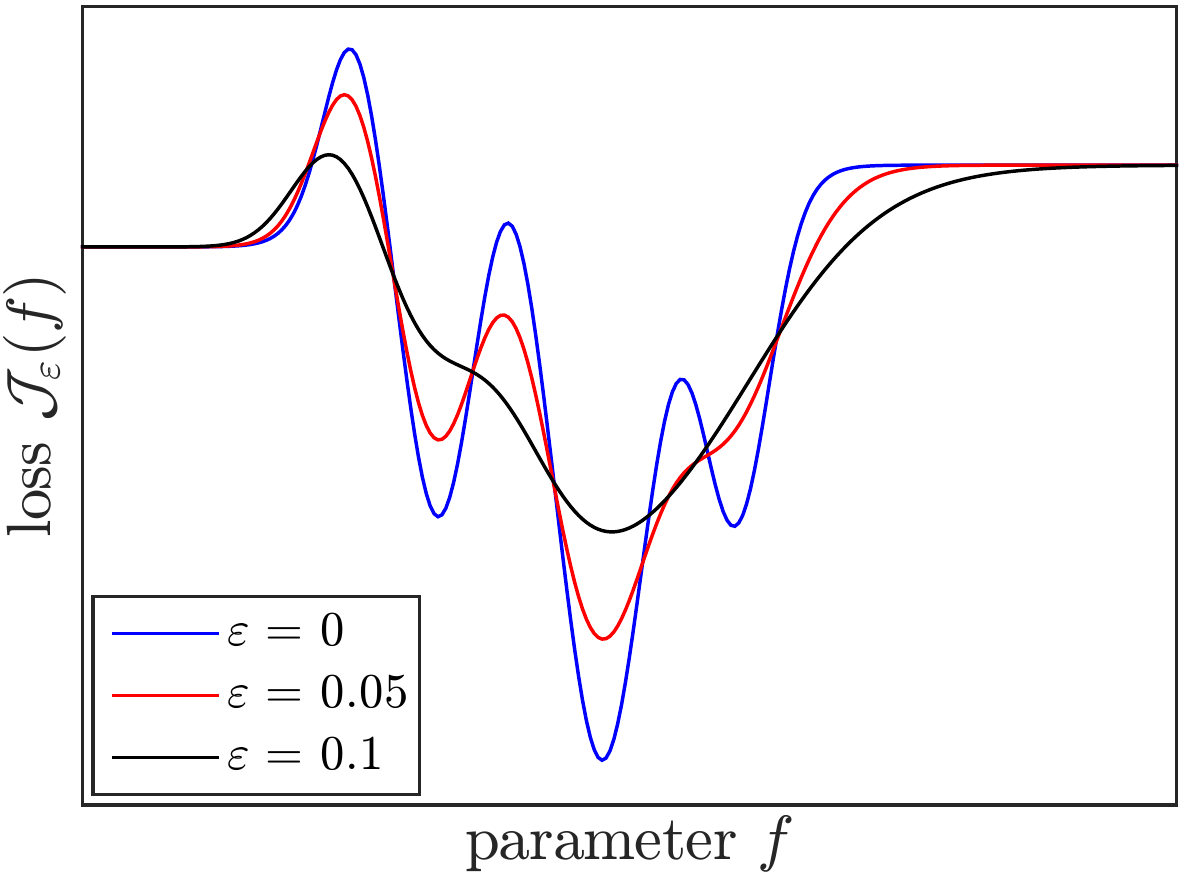}
        \caption{Regularization effect w.r.t. $\ell_1$-loss}
        \label{Fig-l1-loss}
    \end{subfigure}
    ~    
    \begin{subfigure}[t]{0.45\textwidth}
        \centering
        \includegraphics[width=0.9\textwidth]{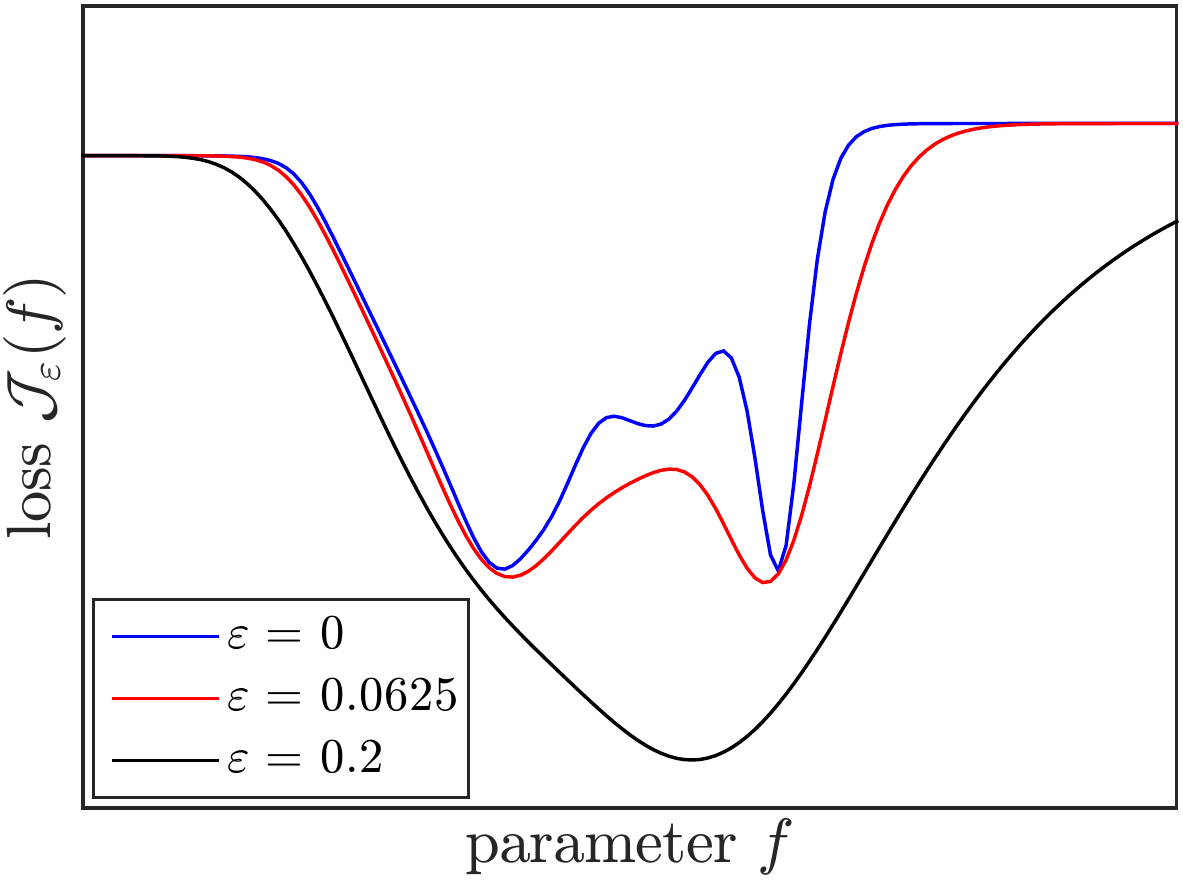}
        \caption{Over-damping effect w.r.t. $\ell_2$-loss}
        \label{Fig-Oversmoothing}
    \end{subfigure}%
    \caption{Left: Similar regularization effects are captured for the $\ell_1$-loss. Right: If too much regularization, or damping, is imposed, the trained model would fail to fit the given dataset.}     
    \label{Fig-L1-loss-const}
\end{figure*}

To begin with, we consider a binary classification problem in which the model parameter or, equivalently, the velocity function $f(x,t)$ is a constant value to be determined, and the operation \eqref{ResNet-Architecture-First-Layer} is set to be an identity mapping $\mathcal{S}=\textnormal{id}$. Note that for this concrete example, there is one redundant component in the terminal condition \eqref{Binary-Classification-Prob-Terminal-Condition}. As such, we may, without any loss of generality, assume that $\mathcal{T}(x)=\sigma(x)=\frac{1}{1+e^{-x}}$ is the sigmoid function and $\Omega=\mathbb{R}$.

Accordingly, for the plain ResNet \eqref{Opt-Ctl-PDE-Plain-ResNet}, the solution to constrained equations \eqref{Backward-Transport-Equation} and the loss function \eqref{Loss-Function-Plain-ResNet} are, respectively, given by
\begin{equation*}
	v(x,t)=\sigma\left(x+(1-t)f\right)\ \ \textnormal{and}\ \ \mathcal{J}_0(f) = \mathbb{E}_{\,y\in\mathbb{R}}\Big[ \left \| \sigma(y+f)-h(y) \right \| \Big].
\end{equation*}
On the contrary, by reversing time $\tilde{u}(x,t)=u(x,1-t)$, the backward Kolmogorov's equations \eqref{Backward-Kolmogorov-Equation} can be reformulated as an initial value problem in forward time, that is,
\begingroup
\renewcommand*{\arraystretch}{1.5}
\begin{equation}
\left\{
\begin{array}{ll}
\displaystyle \tilde{u}_t - f\cdot \tilde{u}_x - \frac{\varepsilon^2}{2}\, f^2\cdot\tilde{u}_{xx} =0\ \  &\textnormal{in}\ \, \mathbb{R}\times (0,1], \\
\tilde{u}(x,0) = \sigma(x)\ \ &\textnormal{in}\ \,\mathbb{R},
\end{array}\right.
\label{Forward-Kolmogorov-Equation-1D}
\end{equation}
\endgroup
where $g(x,t)=f\in\mathbb{R}$. By the Fourier transform \cite{evans2010partial}, the solution to constraint \eqref{Forward-Kolmogorov-Equation-1D} takes the form
\begin{equation*}
	\tilde{u}(x,t) = \sigma\ast\varphi\, (x,t)\ \ \textnormal{where}\ \ \varphi(x,t) = \frac{1}{\sqrt{2\pi t\varepsilon^2f^2}}\ \textnormal{exp} \left(-\frac{(x+tf)^2}{2t\varepsilon^2f^2}\right),
\end{equation*}
and the associated loss function \eqref{Loss-Function-Dropout-ResNet} is thus defined as
\begin{equation*}
	\mathcal{J}_\varepsilon(f) = \mathbb{E}_{\,y\in\mathbb{R}}\Big[ \left \| \sigma\ast\varphi(y,1)-h(y) \right \| \Big].
\end{equation*}

Given a particular dataset consisting of the sample $y\in\mathbb{R}$ and its ground-truth label $h(y)$ $\in\mathbb{Z}_2$, \autoref{Fig-L2-loss-const} implies that the non-convex and rugged landscape for quadratic loss, namely, $\lVert \cdot \rVert = \lVert \cdot \rVert_{2}$ in both \eqref{Loss-Function-Plain-ResNet} and \eqref{Loss-Function-Dropout-ResNet}, can be flattened out through stochastic training, which offers the capability of escaping from poor local minima and therefore forces the optimization algorithms toward better optima. More specifically, the energy barriers between pairs of local minima can be lessened by injecting noise so that the convexity of rugged loss landscape is enhanced to aid optimization. Similar regularization effects are observed in \autoref{Fig-L1-loss-const} for the $\ell_1$-loss (\textit{i.e.}, $\lVert \cdot \rVert = \lVert \cdot \rVert_{1}$), which validate the effectiveness of our method.

However, injecting more noise in the feature flow \eqref{Dropout-ResNet-Continuous-DataFlow} makes it harder to fit the training data. As a consequence, the resulting model parameters could be invalid for inference as depicted in \autoref{Fig-Oversmoothing}, since the effective information in the flow of feature map is swamped by the injected noise. Put differently, stochastic training of a plain ResNet incurs a trade-off between model regularization and data fitting, where both parts are imperative to improve the generalization performance. Therefore, instead of simply setting the survival probability to be the default value $(p=0.5)$ \cite{srivastava2014dropout} that may cause negative feedback \cite{he2016identity}, we argue here that a fine-tuned noise level can improve the generalization capability of trained models, and experimental results in the following section validate our statements.

\begin{figure}[htp]

	\centering	
    \begin{subfigure}[t]{0.45\textwidth}
        \centering
        \includegraphics[width=0.98\textwidth]{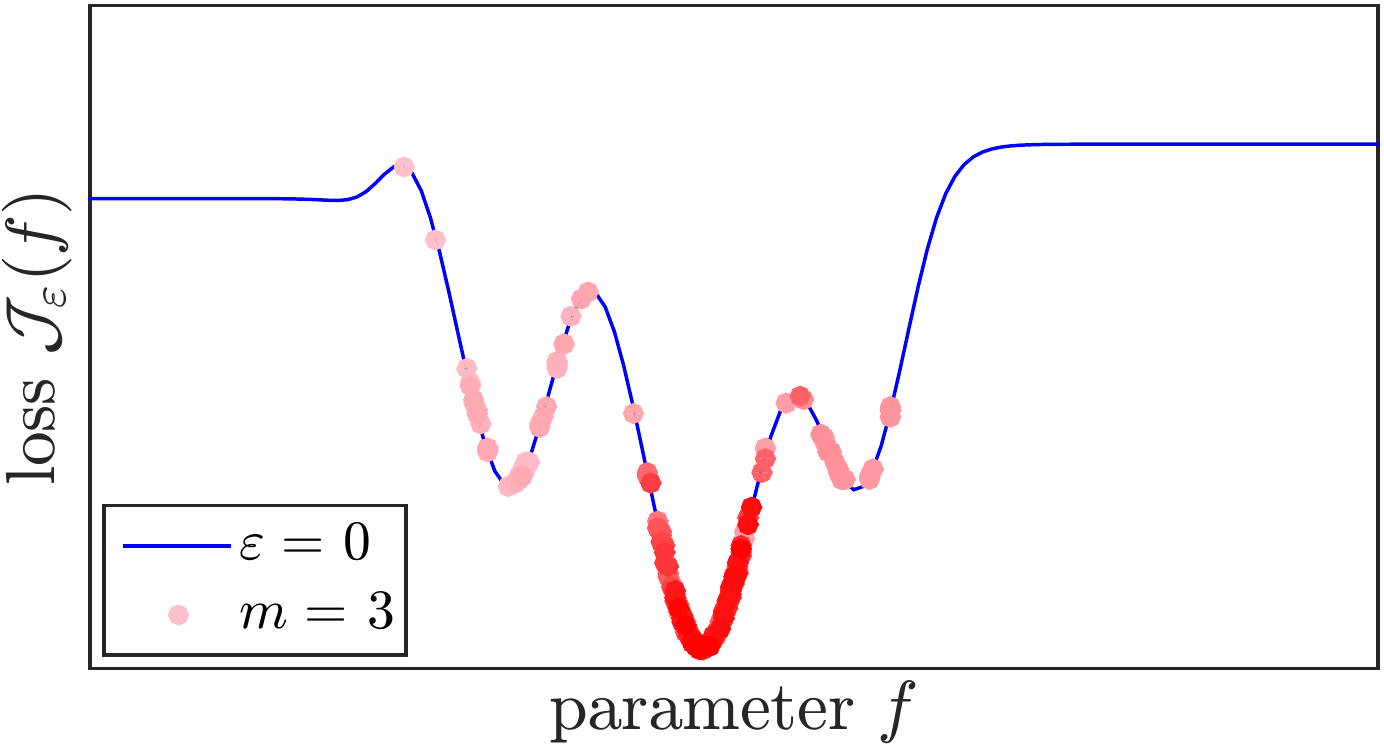}
        \caption{Small-batch SGD on a plain loss}
        \label{Fig-SB-PlainLoss-Case1}
    \end{subfigure}%
    ~
    \begin{subfigure}[t]{0.45\textwidth}
        \centering
        \includegraphics[width=0.98\textwidth]{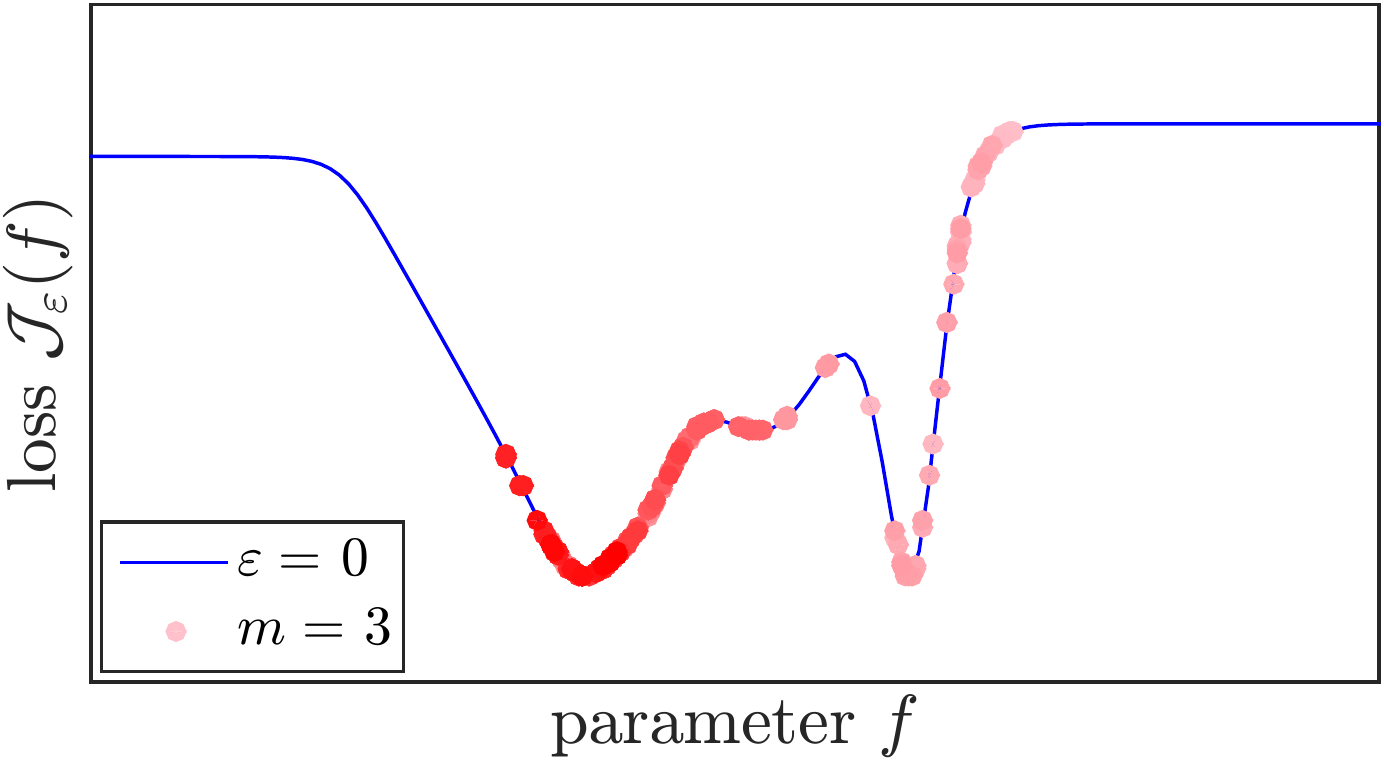}
        \caption{Small-batch SGD on a plain loss}
        \label{Fig-SB-PlainLoss-Case2}
    \end{subfigure}%

    \begin{subfigure}[t]{0.45\textwidth}
        \centering
        \includegraphics[width=0.98\textwidth]{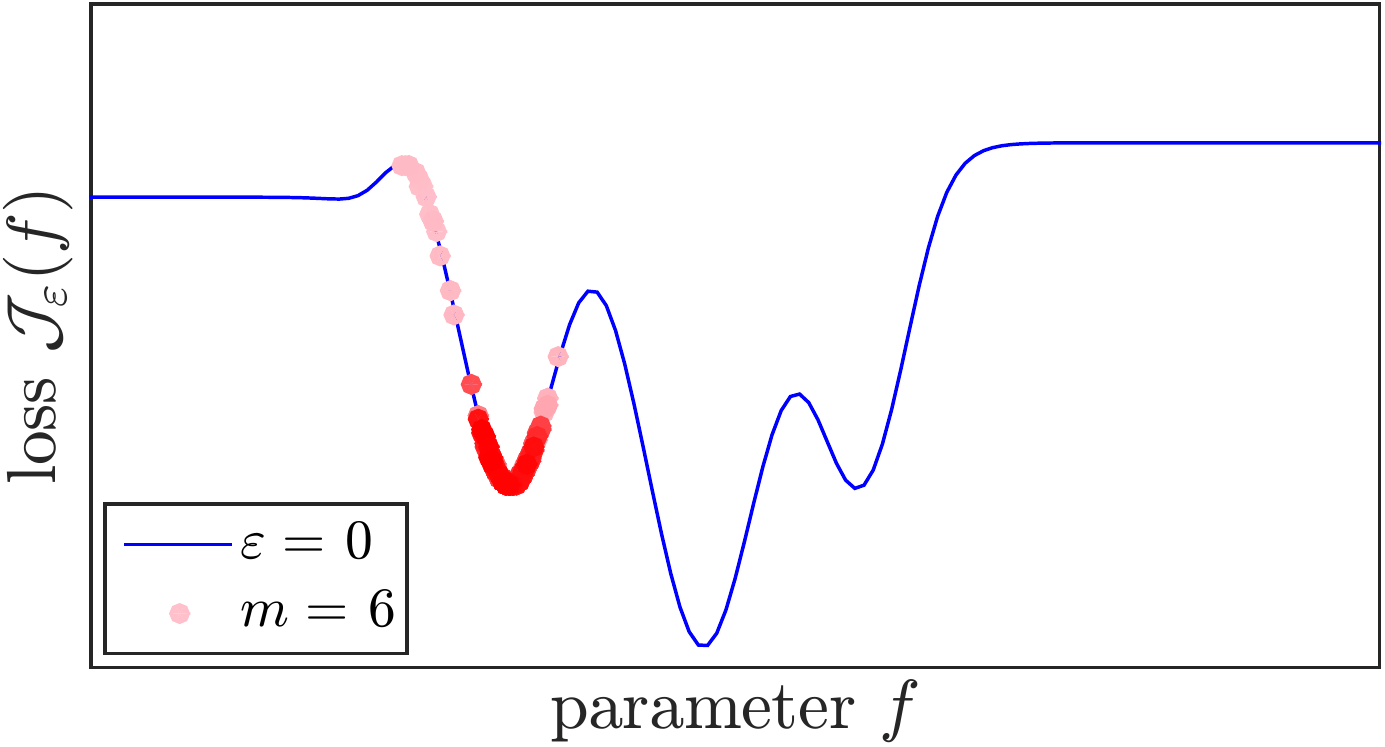}
        \caption{Large-batch SGD on a plain loss}
        \label{Fig-LB-PlainLoss-Case1}
    \end{subfigure}%
    ~
    \begin{subfigure}[t]{0.45\textwidth}
        \centering
        \includegraphics[width=0.98\textwidth]{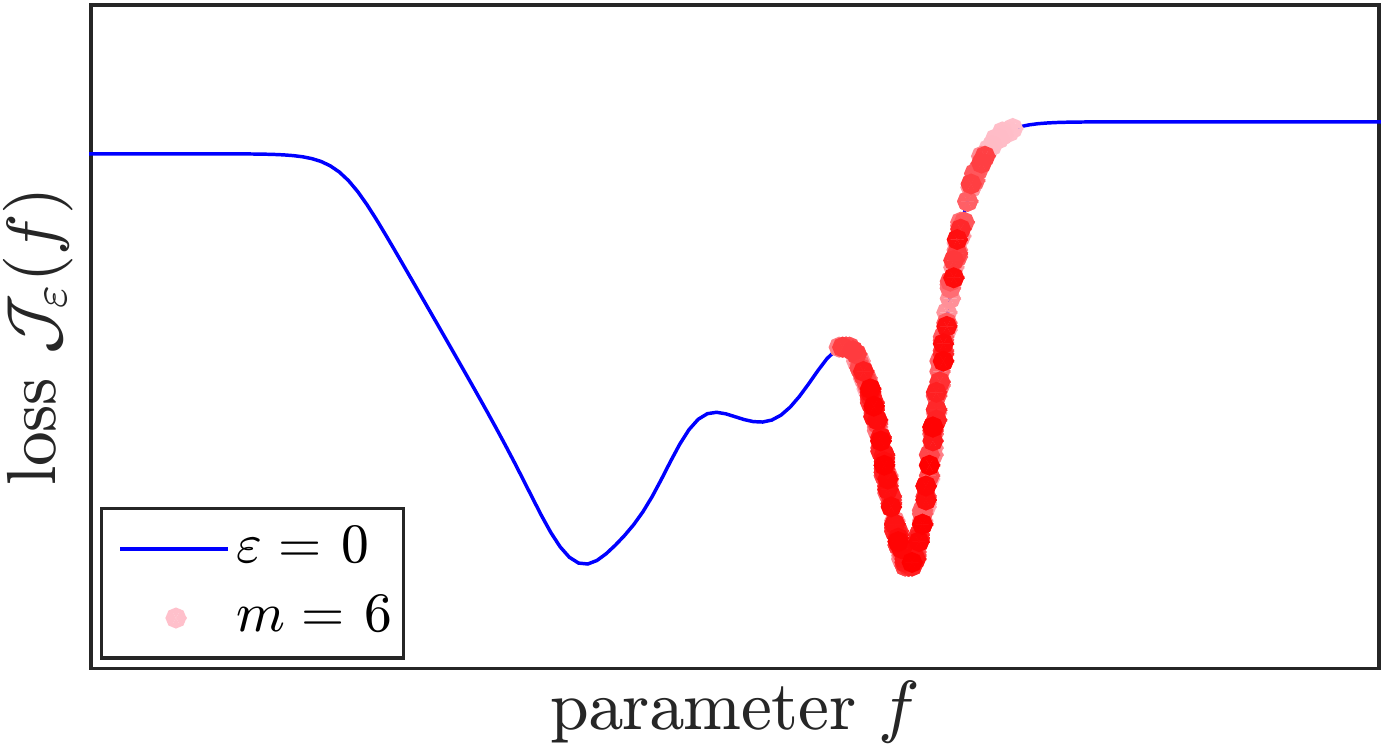}
        \caption{Large-batch SGD on a plain loss}
        \label{Fig-LB-PlainLoss-Case2}
    \end{subfigure}%

    \begin{subfigure}[t]{0.45\textwidth}
        \centering
        \includegraphics[width=0.98\textwidth]{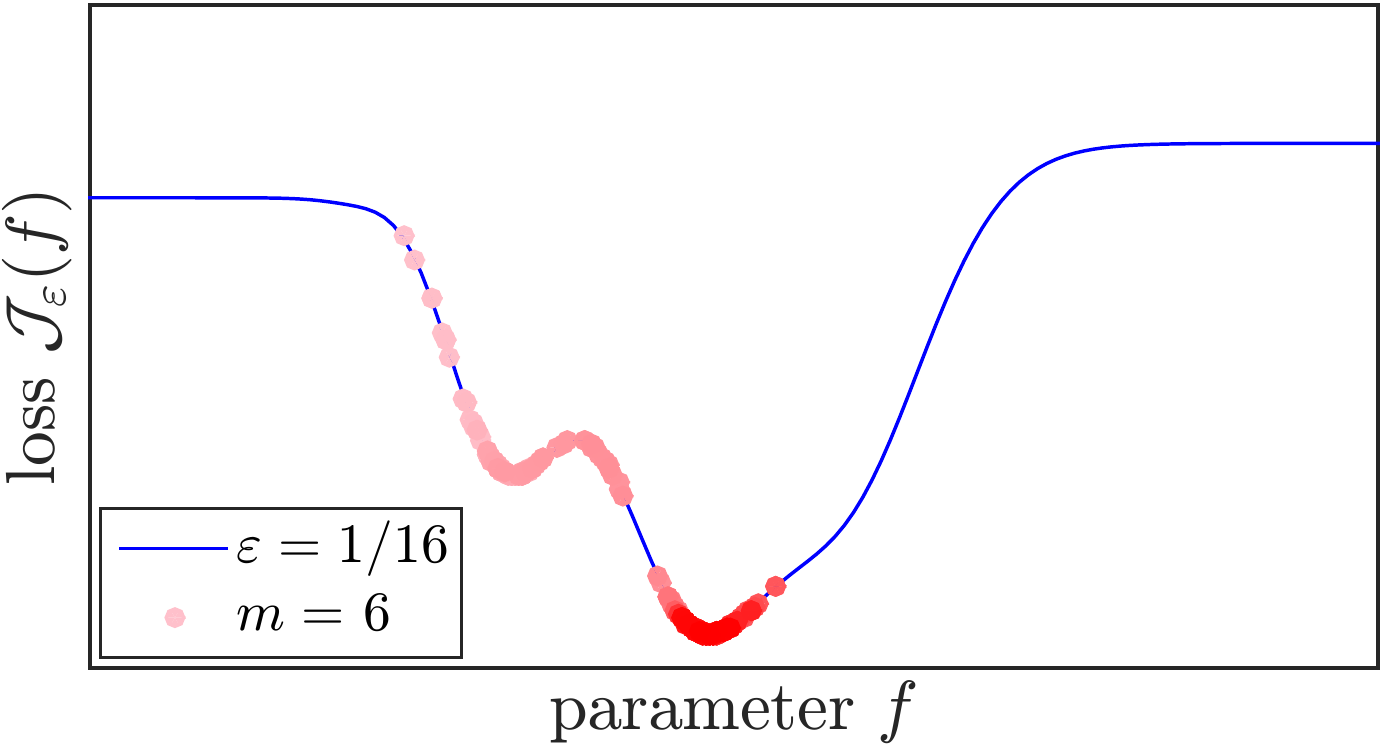}
        \caption{Large-batch SGD on a regularized loss}
        \label{Fig-LB-RegularizedLoss-Case1}
    \end{subfigure}%
    ~
    \begin{subfigure}[t]{0.45\textwidth}
        \centering
        \includegraphics[width=0.98\textwidth]{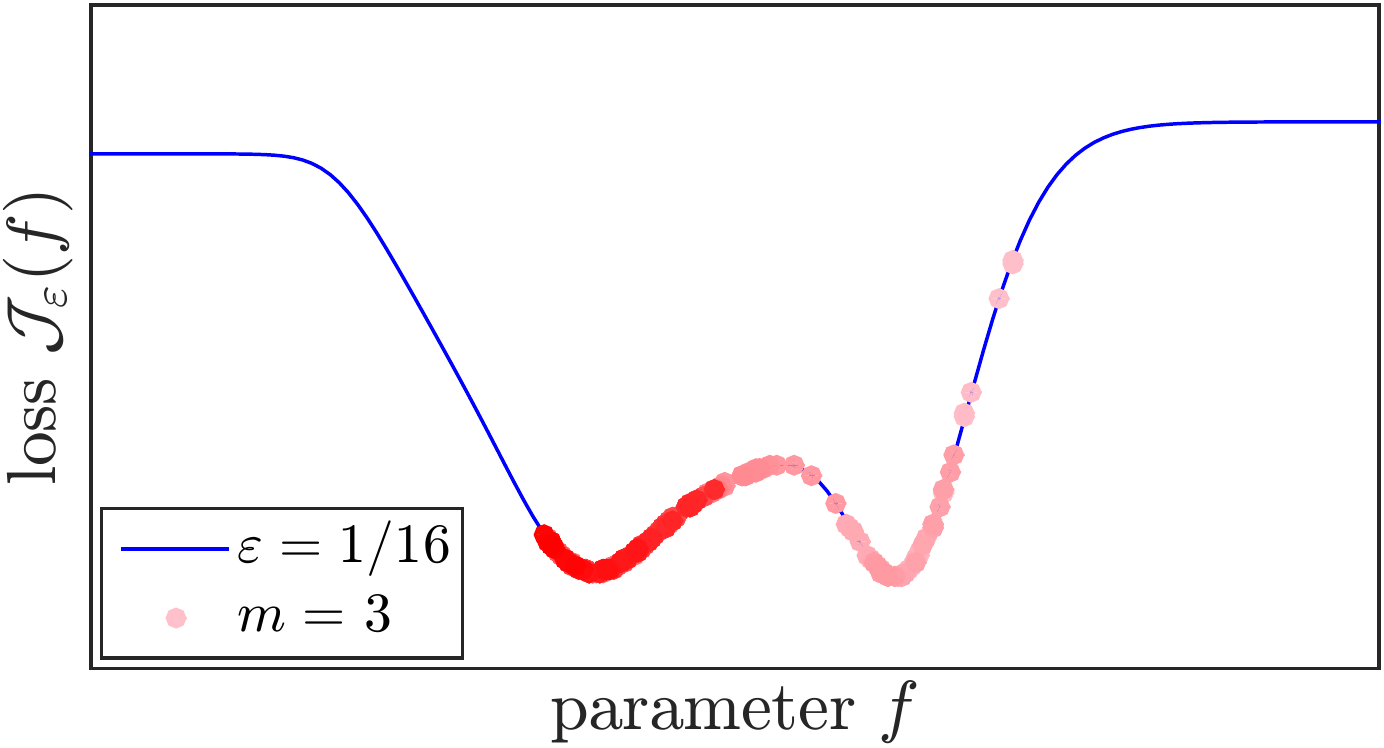}
        \caption{Large-batch SGD on a regularized loss}
        \label{Fig-LB-RegularizedLoss-Case2}
    \end{subfigure}%
    
    \begin{subfigure}[t]{0.45\textwidth}
        \centering
        \includegraphics[width=0.98\textwidth]{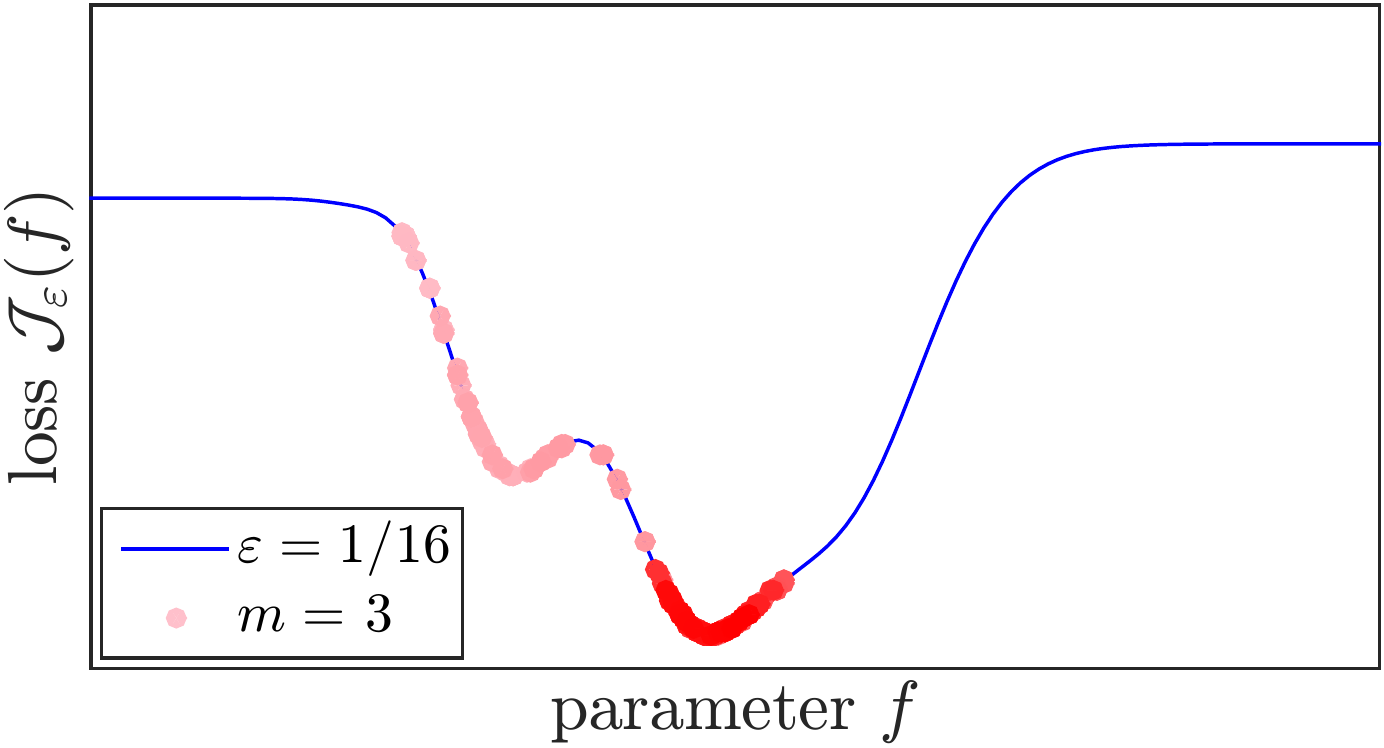}
        \caption{Small-batch SGD on a regularized loss}
        \label{Fig-SB-RegularizedLoss-Case1}
    \end{subfigure}%
    ~
    \begin{subfigure}[t]{0.45\textwidth}
        \centering
        \includegraphics[width=0.98\textwidth]{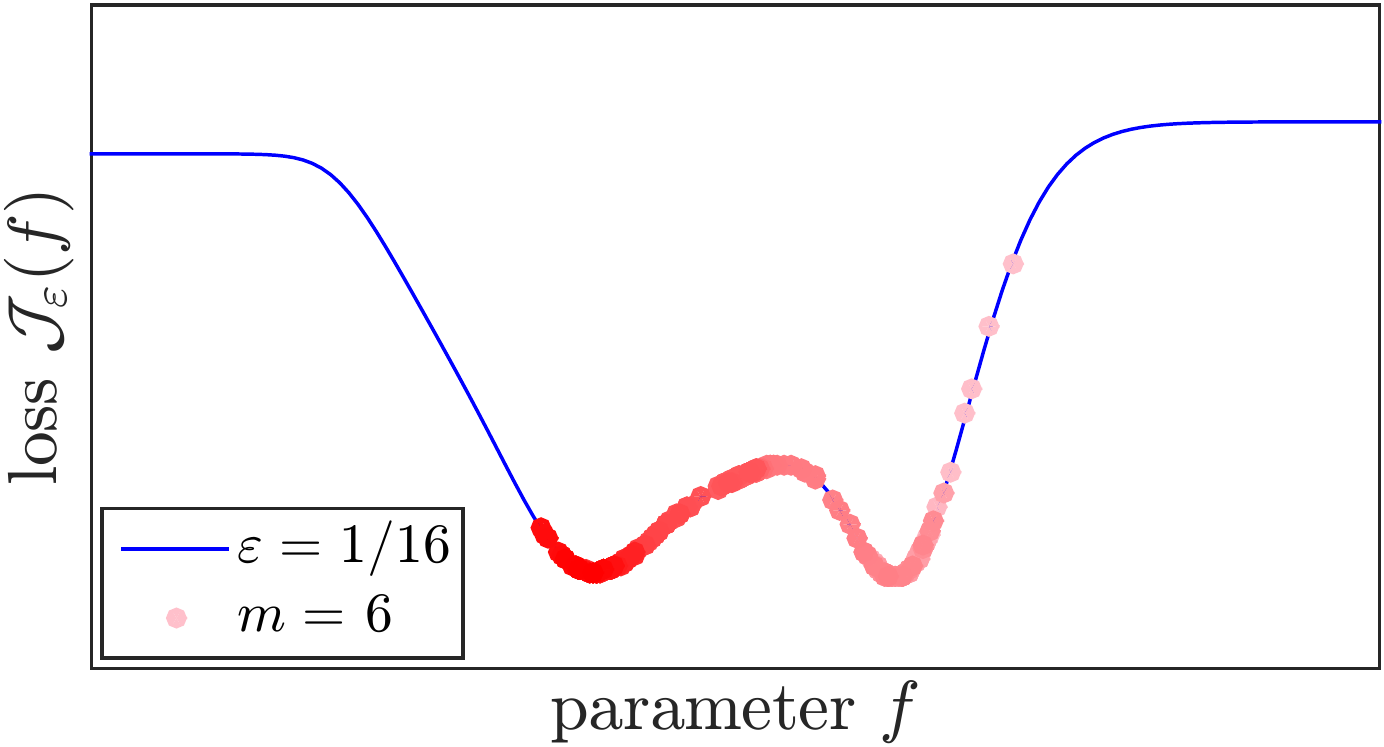}
        \caption{Small-batch SGD on a regularized loss}
        \label{Fig-SB-RegularizedLoss-Case2}
    \end{subfigure}%
       
    \caption{Small/large-batch SGD iterative solutions on the plain/regularized $\ell_2$-loss landscape where the color is arranged from light pink to dark red as the iteration increases. The left and right columns correspond to two different datasets, and the same setting is utilized for each column except for the noise level $\varepsilon$ and batch-size $m$.}
    \label{Fig-L2-Loss-Const-SGD-Solu}
\end{figure}

Since the machine learning models are typically trained using SGD or one of its variants, we show, in \autoref{Fig-L2-Loss-Const-SGD-Solu}, the SGD iterative solutions on a plain/regularized loss landscape using a small/large-batch size $m\in\mathbb{Z}_+$ (not relabeled). As mentioned before, the loss landscape of plain ResNet is highly rugged as depicted in \autoref{Fig-LB-PlainLoss-Case1}, where the large-batch SGD approximations are trapped in sharp ravine due to the high energy barrier between pairs of local minima. Although this issue can be circumvented by using a small-batch method as shown in \autoref{Fig-SB-PlainLoss-Case1}, large-batch methods are preferable in practice due to computing efficiency \cite{smith2017don}. Hence, the stochastic training strategy is highly desirable for practical applications since the training task can be well simplified due to the smoothing effects (see \autoref{Fig-SB-RegularizedLoss-Case1} and \autoref{Fig-LB-RegularizedLoss-Case1} for example).

In addition, recent work by \cite{chaudhari2016entropy,keskar2016large,hoffer2017train,hu2017diffusion} conjectures that the flatness of minimizer in the loss landscape is critically related to the generalization error of trained networks. That is, the optima found by small-batch methods usually lie in relatively flat ravines and are more likely to have a small gap between error on the training and test datasets, even if the training loss is worse than for the sharp minima found by large-batch methods. As such, we consider a double well like potential depicted in \autoref{Fig-LB-PlainLoss-Case2} and \autoref{Fig-SB-PlainLoss-Case2}, where the stochastic regularization effects shown in \autoref{Fig-LB-RegularizedLoss-Case2} and \autoref{Fig-SB-RegularizedLoss-Case2} allow the SGD algorithm to converge to broader minimizer without tuning the batch size.

\subsection{Experiments on Real-world Dataset}
In this section, we evaluate the proposed stochastic training strategy \eqref{Dropout-ResNet-Discrete-DataFlow} for a real-world image classification task, \textit{i.e.}, the CIFAR-10 dataset \cite{krizhevsky2009learning} where the dropout technique has been actively used \cite{huang2016deep,gastaldi2017shake,noh2017regularizing}. To isolate the stochastic regularization effects, we compare the empirical performance of PreResNets with or without injecting dropout layers.

\subsubsection{Implementation Details}
Models are trained on the 50k training images with a batch size of $128$, a weight decay of $10^{-4}$ and momentum $0.9$, which is then evaluated on the 10k test images. Standard data augmentation is utilized \cite{lee2015deeply}. The training starts with a learning rate of $0.1$, and is divided by $10$ at $81$ and $122$ epochs. All models are trained on $2$ Nvidia Tesla GPUs and terminated after $164$ epochs.

During training, the retained activation is scaled up by $1/p$ as shown in \eqref{Dropout-ResNet-Discrete-DataFlow}, where the dropout tensor $\gamma_k$ is overwritten with new random numbers over iterations. At test time, the plain PreResNet \eqref{Plain-ResNet-Discrete-DataFlow} with learned model parameters is used for validation, \textit{i.e.}, no noise is injected in validation or, equivalently, $p\equiv 1$ in \eqref{Dropout-ResNet-Discrete-DataFlow}. To be specific, the proposed training procedure using BasicBlock (see Remark \ref{Remark-BasicBlock}) is defined by
\begin{equation}
\label{Training-Single-Path}
\begin{split}
	& \displaystyle \underset{\mathcal{S},\mathcal{F},\mathcal{T}}{\textnormal{minimize}}\ \ \mathcal{J}_\varepsilon(\mathcal{S},\mathcal{F},\mathcal{T}) = \frac{1}{|\Omega|}\sum_{y_i\in\Omega} \left\|\,  \mathcal{T}(X_K\,|\,X_0=\mathcal{S}y_i ) - h(y_i)\, \right\|,\\
	& \textnormal{subject to}\ \ X_{k+1} = X_k + \Big[ w_k^{(2)}\cdot a\left(w_k^{(1)}\cdot a(X_k)\right) \Big]  \odot \frac{\gamma_k}{p},
\end{split}
\end{equation}
where $\mathcal{J}_\varepsilon$ is the empirical loss (not relabeled) over the training database $\left\{y_i,h(y_i)\right\}_{i=1}^{50k}$, $\gamma_k\sim\textnormal{Bern}(p)$ the dropout mask tensor where $p\in(0,1]$ during training and $p=1$ in test mode, $\lVert\cdot\rVert$ the cross-entropy loss for multi-class classification task, and $0\leq k \leq K-1$ in which $K$ is the total number of residual blocks.

\subsubsection{Experimental Results}
\begin{table}[t]
\begin{center}
\small
\begin{tabular}{c||cccccccc}
\toprule
$p$ & $85\%$ & $87.5\%$ & $90\%$ & $92.5\%$ & $95\%$ & $97.5\%$ & $100\%$ \\
\midrule
\multirow{2}{*}{PreResNet-20} & \multirow{2}{*}{7.65} & \multirow{2}{*}{7.67} & \textbf{7.59} & \multirow{2}{*}{7.80} & \multirow{2}{*}{7.91} &  \multirow{2}{*}{7.98} &   8.22 \\
& & & \textbf{(7.28)}                      &       &  &  &   (7.98) \\
\midrule
\multirow{2}{*}{PreResNet-56} & \textbf{6.19} & \multirow{2}{*}{6.24} & \multirow{2}{*}{6.22} & \multirow{2}{*}{6.28}   & \multirow{2}{*}{6.30} &  \multirow{2}{*}{6.43} &  6.55 \\
& \textbf{(6.05)} & &     &   &  &   &   (6.44) \\
\midrule
 \multirow{2}{*}{PreResNet-110} & \multirow{2}{*}{5.88} & \textbf{5.58} & \multirow{2}{*}{5.74}   & \multirow{2}{*}{5.65} & \multirow{2}{*}{5.69} &  \multirow{2}{*}{5.75} &  6.00 \\
&  & \textbf{(5.48)} &   &      &  &  &    (5.86) \\
\bottomrule
\end{tabular}
\end{center}
\caption{Top-1 error rates ($\%$) on CIFAR-10 test set. All the results are mean value of 4 runs where the best results (with the highest accuracy in the bracket) are displayed in bold.}
\label{Table-results-CIFAR10}
\end{table}

\begin{figure*}[t!]
    \centering
    \begin{subfigure}[t]{0.45\textwidth}
        \centering
        \includegraphics[width=0.97\textwidth]{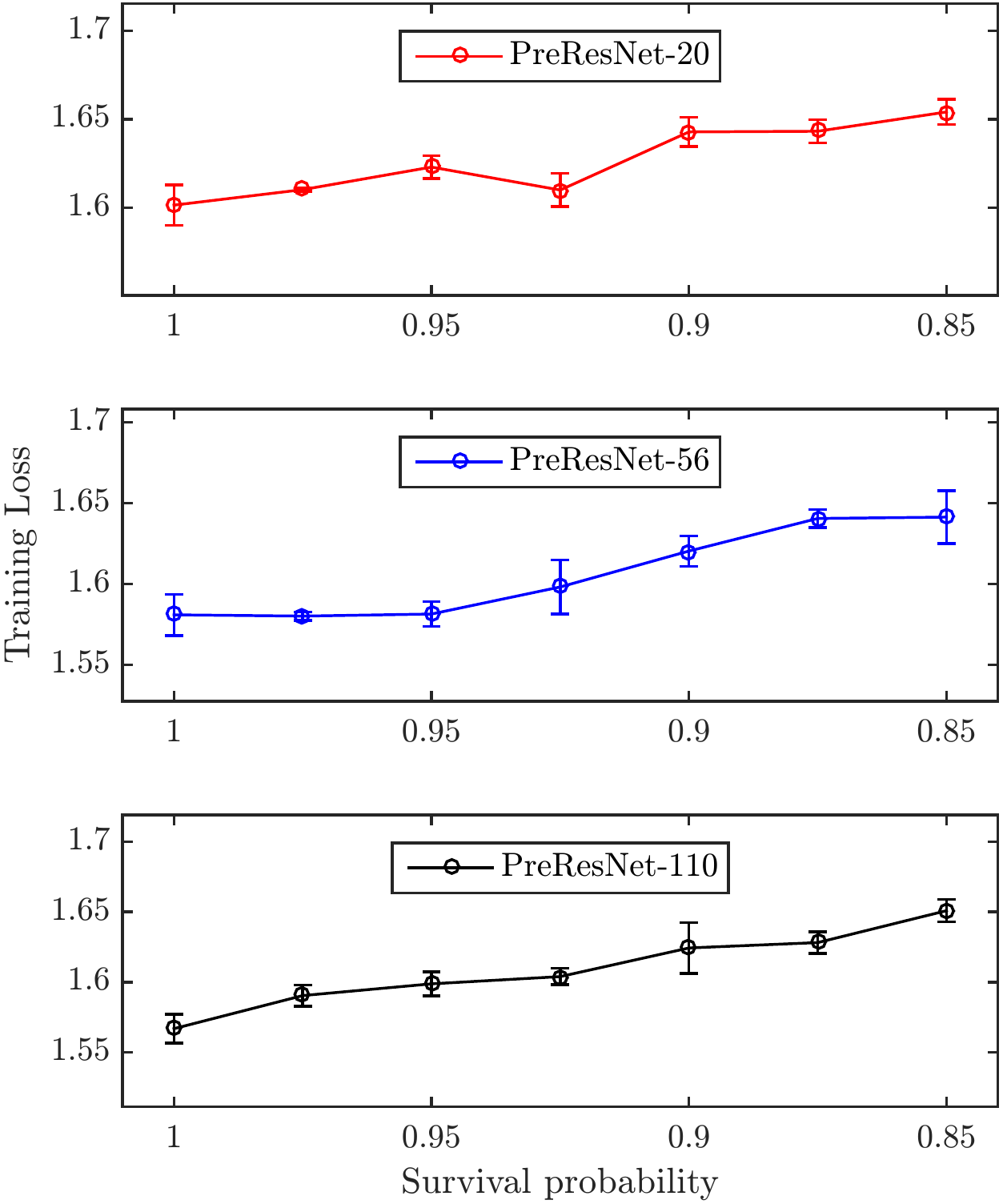}
        \caption{Error bars for training loss}
    \end{subfigure}%
    ~ 
    \begin{subfigure}[t]{0.45\textwidth}
        \centering
        \includegraphics[width=0.97\textwidth]{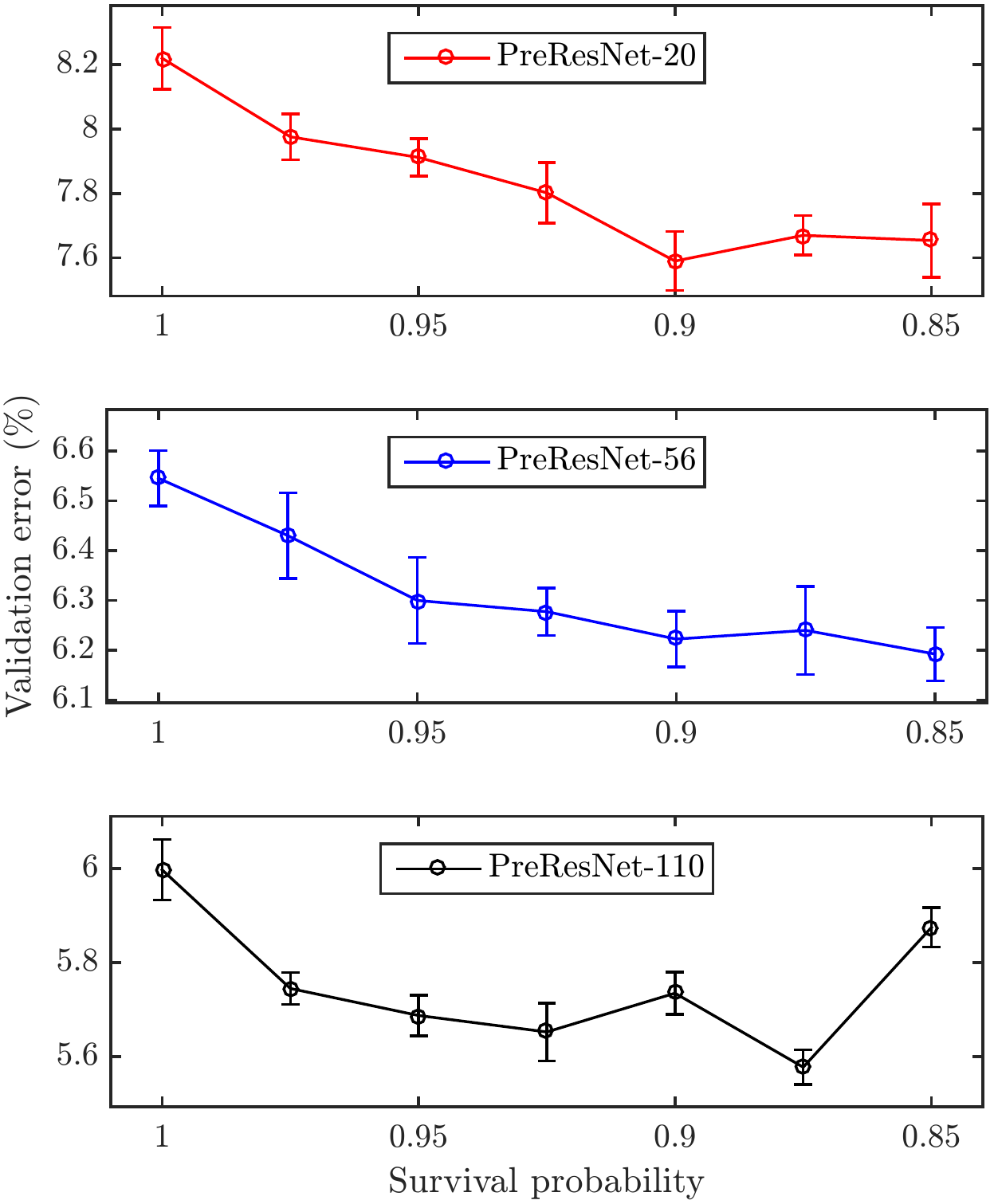}
        \caption{Error bars for validation error}
    \end{subfigure}
    \caption{Training loss and validation error versus survival probability. The results are computed with 4 runs on CIFAR-10, shown with standard error bars.}
   \label{Fig-Error-bars}
\end{figure*}

\begin{figure*}[htp]
    \centering
    \begin{subfigure}[t]{0.45\textwidth}
        \centering
        \includegraphics[width=0.98\textwidth]{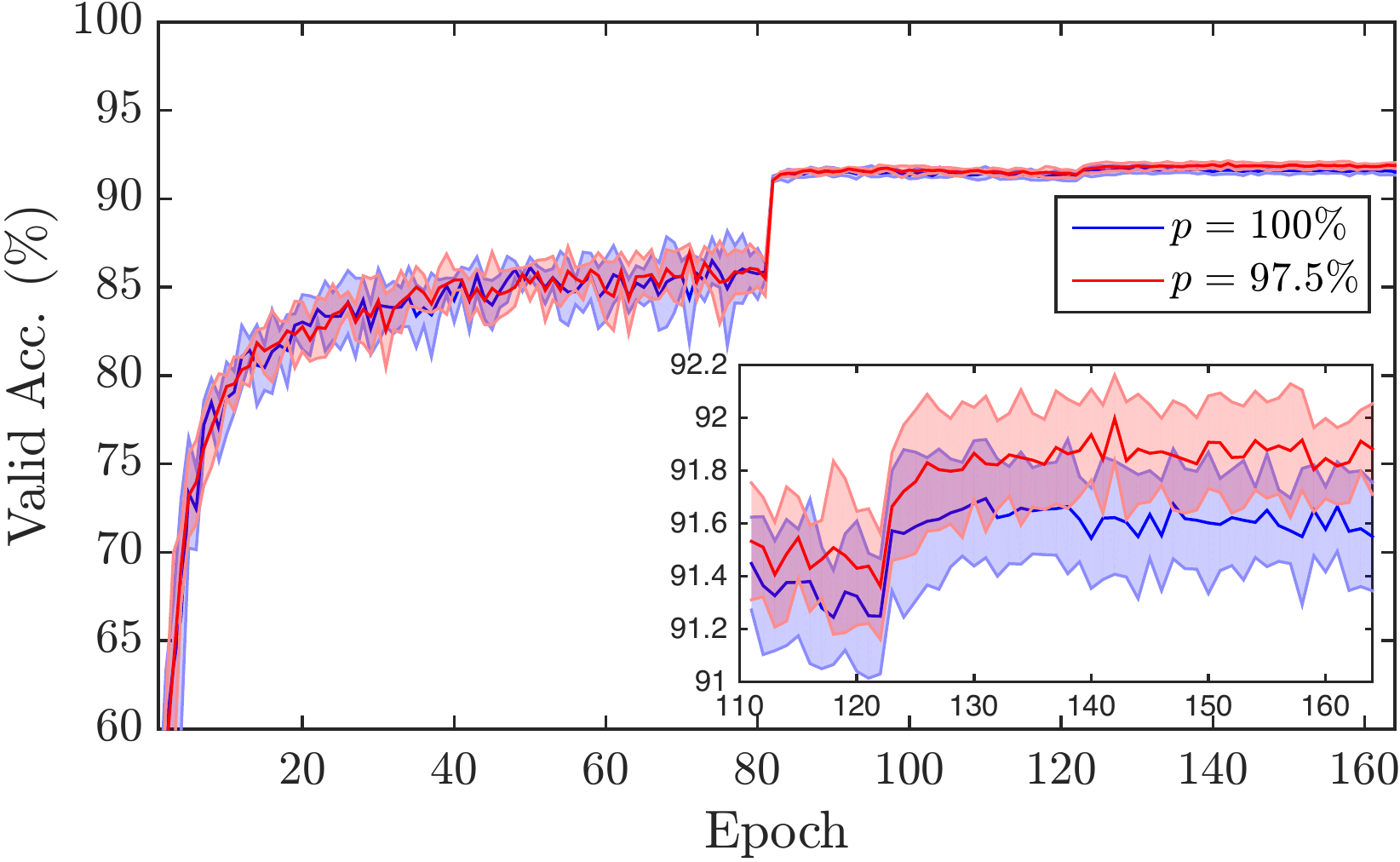}
        \caption{PreResNet-20 with $p=97.5\%$}
    \end{subfigure}%
    ~ 
    \begin{subfigure}[t]{0.45\textwidth}
        \centering
        \includegraphics[width=0.98\textwidth]{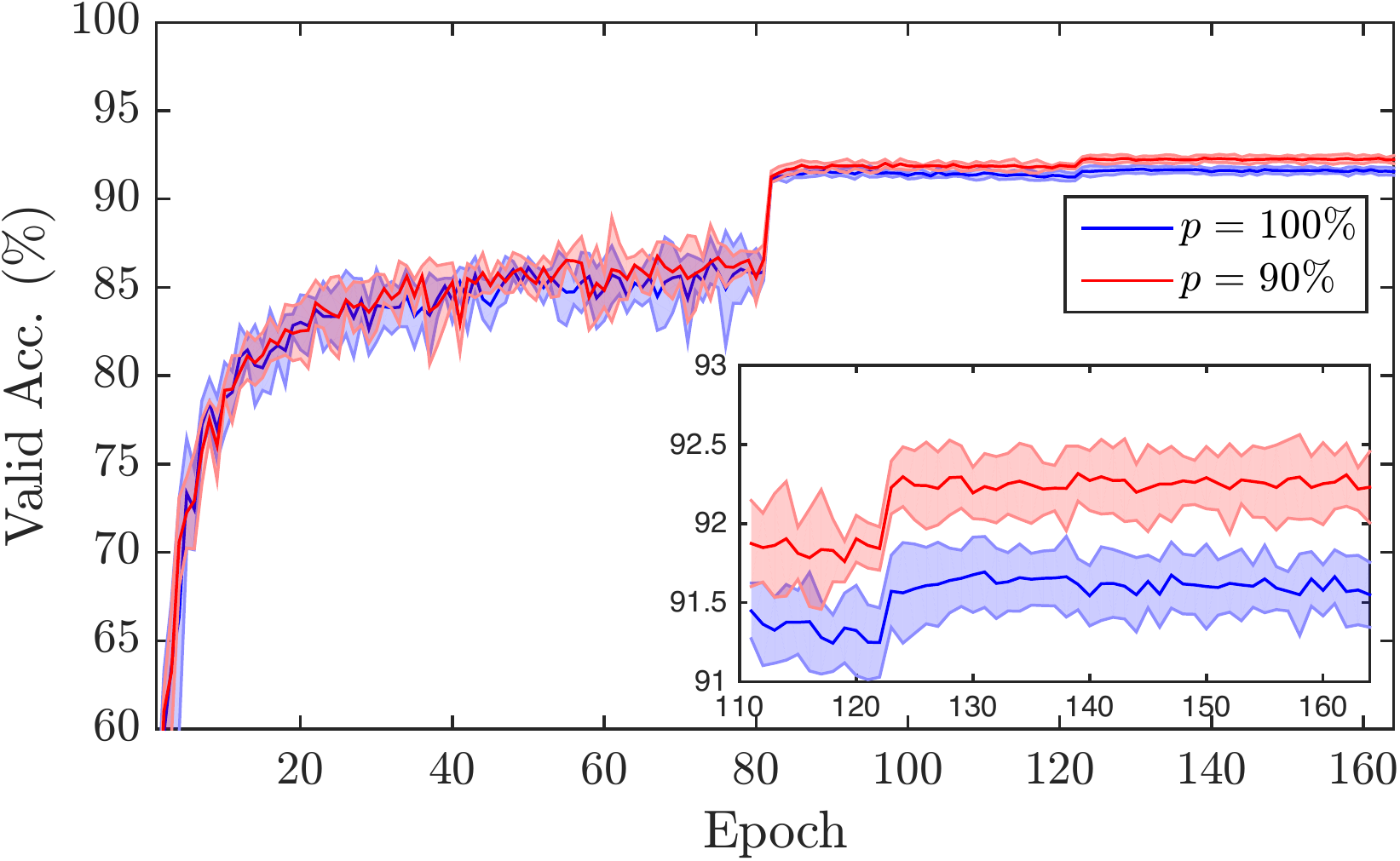}
        \caption{PreResNet-20 with $p=90\%$}
    \end{subfigure}

    \begin{subfigure}[t]{0.45\textwidth}
        \centering
        \includegraphics[width=0.98\textwidth]{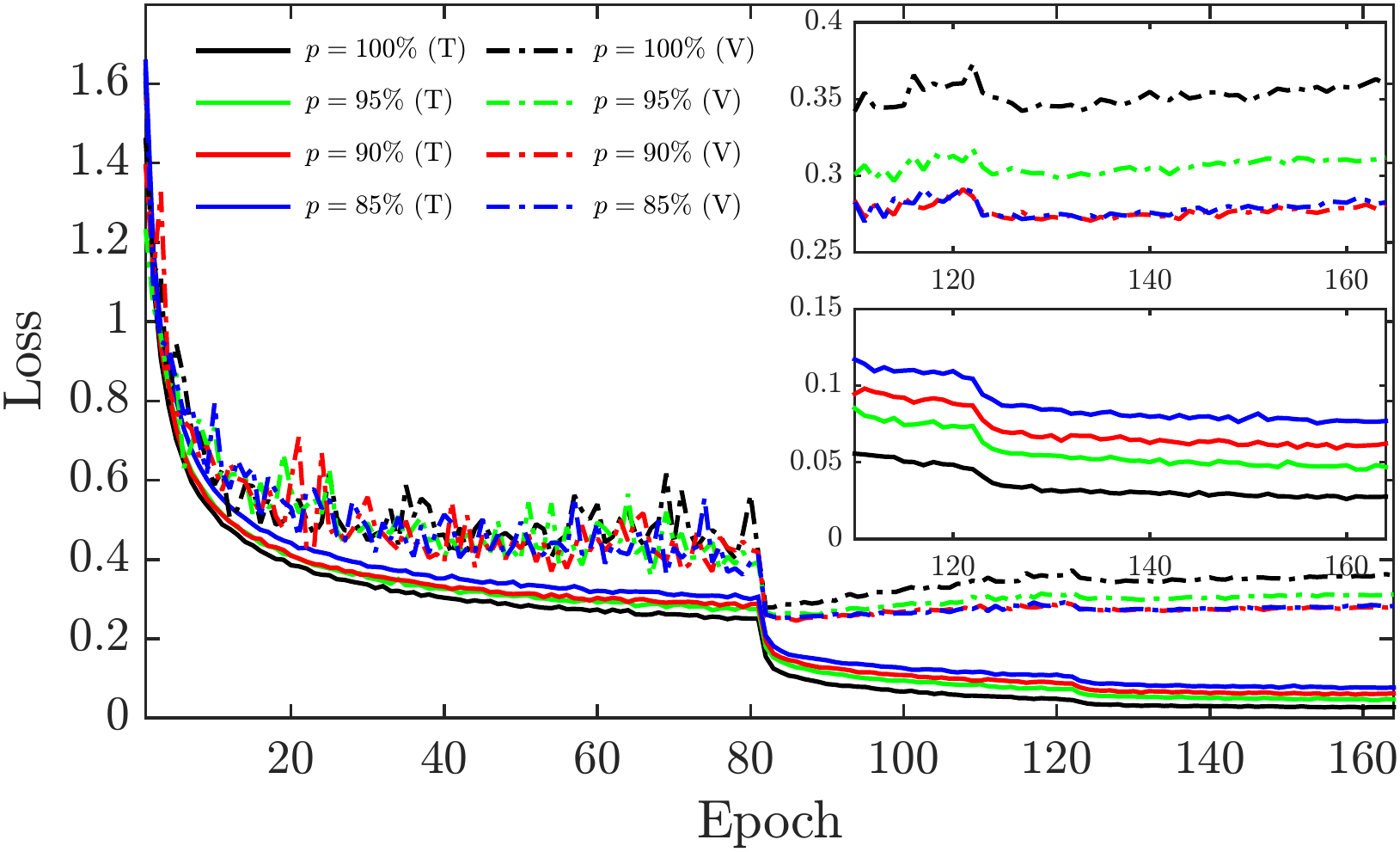}
        \caption{Loss value for PreResNet-20}
    \end{subfigure}%
    ~ 
    \begin{subfigure}[t]{0.45\textwidth}
        \centering
        \includegraphics[width=0.98\textwidth]{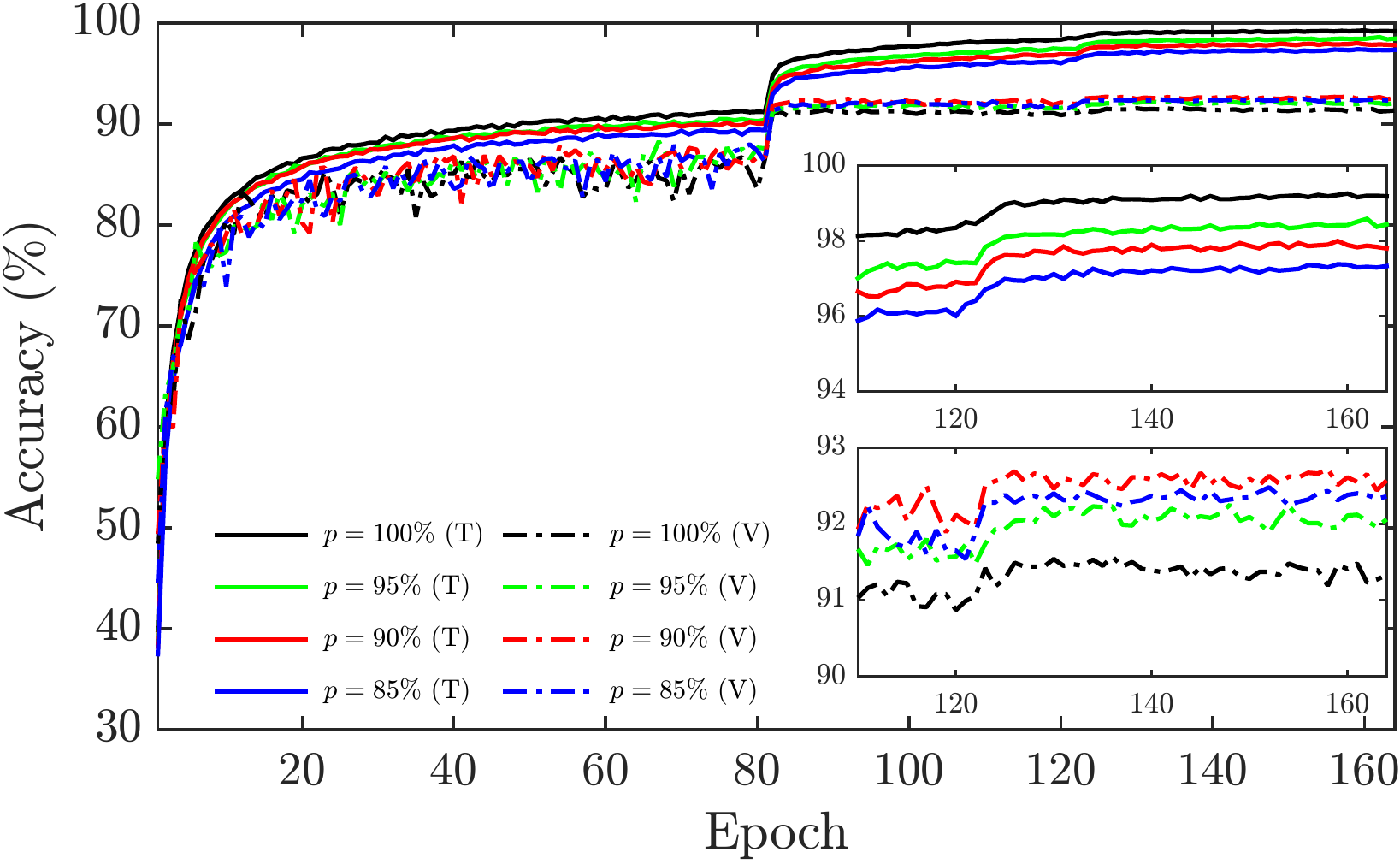}
        \caption{Accuracy for PreResNet-20}
    \end{subfigure}
    \caption{Top: Learning curves (mean $\pm$ variance) for PreResNet-20 with different dropout ratios. Bottom: Learning curves for PreResNet-20 at training (T) and validation (V) during one simulation.}
    \label{Fig-R20-Sample-Path}
           
\end{figure*}

The results in \autoref{Table-results-CIFAR10} represent the validation errors of the proposed PreResNet with a different number of layers and survival probabilities, where the case $p=100\%$ corresponds to the plain PreResNet (baseline). We run each method 4 times and report the mean validation error. The best results, as well as the highest accuracy, are displayed in bold. Graphs showing the standard error bars ($\textnormal{mean}\,\pm$ standard deviation) are given in \autoref{Fig-Error-bars}. Clearly, both \autoref{Table-results-CIFAR10} and \autoref{Fig-Error-bars} imply that the stochastic training strategy \eqref{Training-Single-Path} can improve the generalization capability of trained networks, provided that a suitable survival probability $p$ is used. Otherwise, the neural network may fail to converge to a good solution when eliminating too many connections \cite{he2016identity}, since the loss landscape is somewhat over-damped as the regularization parameter $\varepsilon$ increases.

Moreover, for the learning curve of PreResNet-20, the deviation from its mean is depicted in \autoref{Fig-R20-Sample-Path} where two different dropout probabilities are chosen. Note that, as $p\to100\%$ (or, equivalently, the parameter $\varepsilon\to0$), the improvement in inference is reduced since the regularization term diminishes. The determination of the optimal dropout probability associated with the depth of ResNet, which depends on the regularization of loss landscape associated with the discrete dynamic, is an interesting open question that needs further study. 

\begin{figure*}[htp]
    \centering
    \begin{subfigure}[t]{0.47\textwidth}
        \centering
        \includegraphics[width=0.98\textwidth]{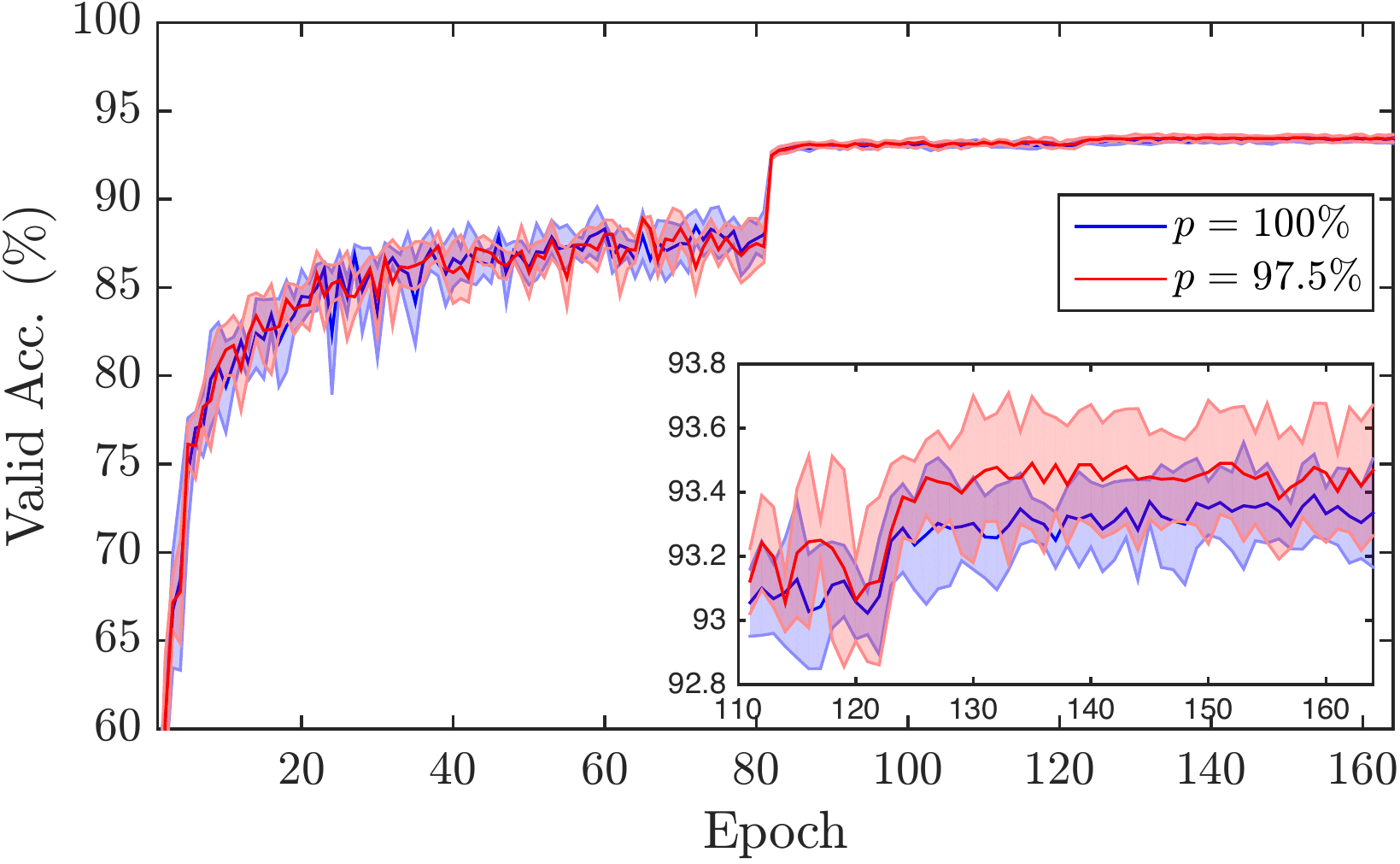}
        \caption{PreResNet-56 with $p=97.5\%$.}
    \end{subfigure}%
    ~ 
    \begin{subfigure}[t]{0.47\textwidth}
        \centering
        \includegraphics[width=0.98\textwidth]{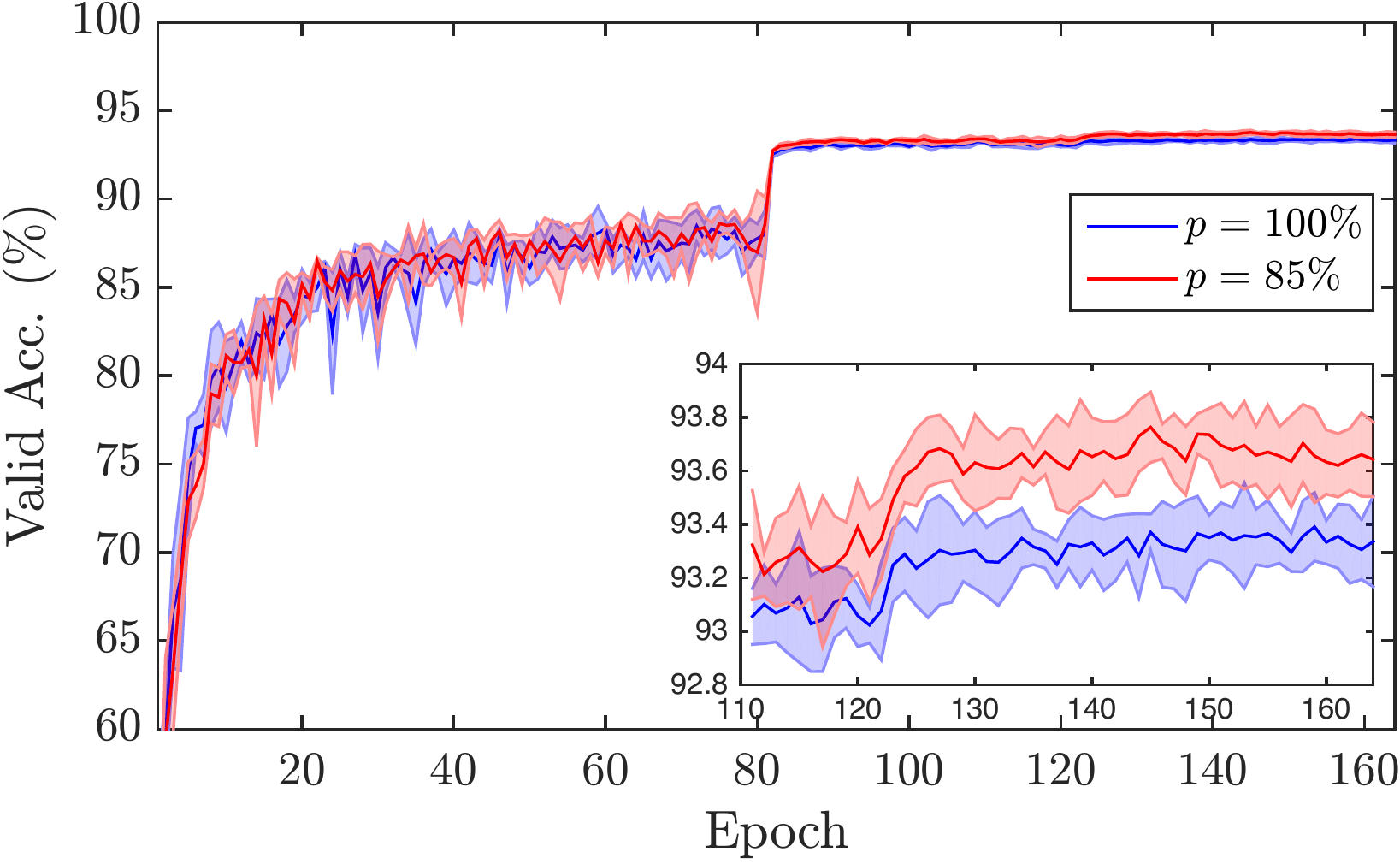}
        \caption{PreResNet-56 with $p=85\%$.}
    \end{subfigure}

    \begin{subfigure}[t]{0.47\textwidth}
        \centering
        \includegraphics[width=0.98\textwidth]{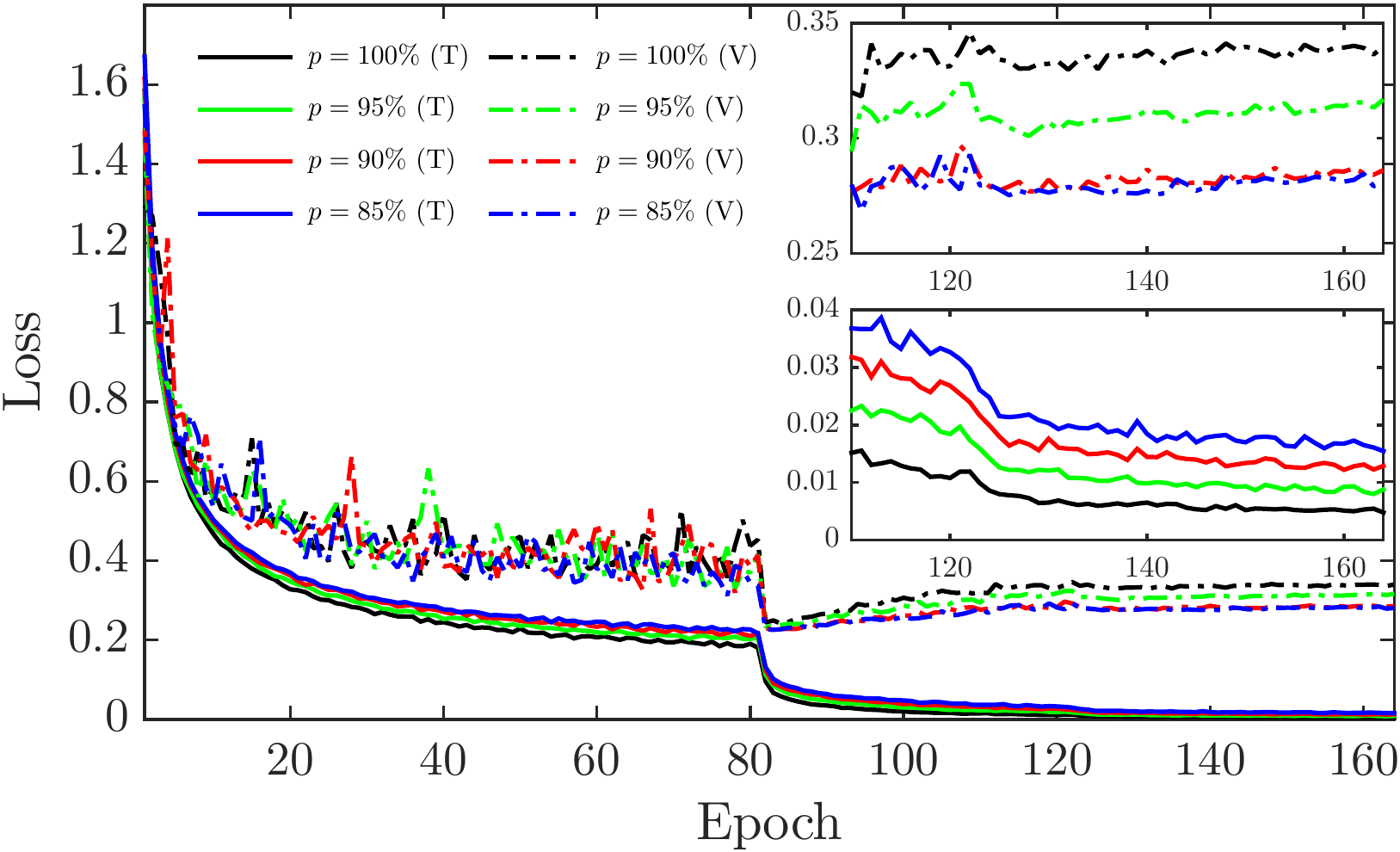}
        \caption{Loss value for PreResNet-56.}
    \end{subfigure}%
    ~ 
    \begin{subfigure}[t]{0.47\textwidth}
        \centering
        \includegraphics[width=0.98\textwidth]{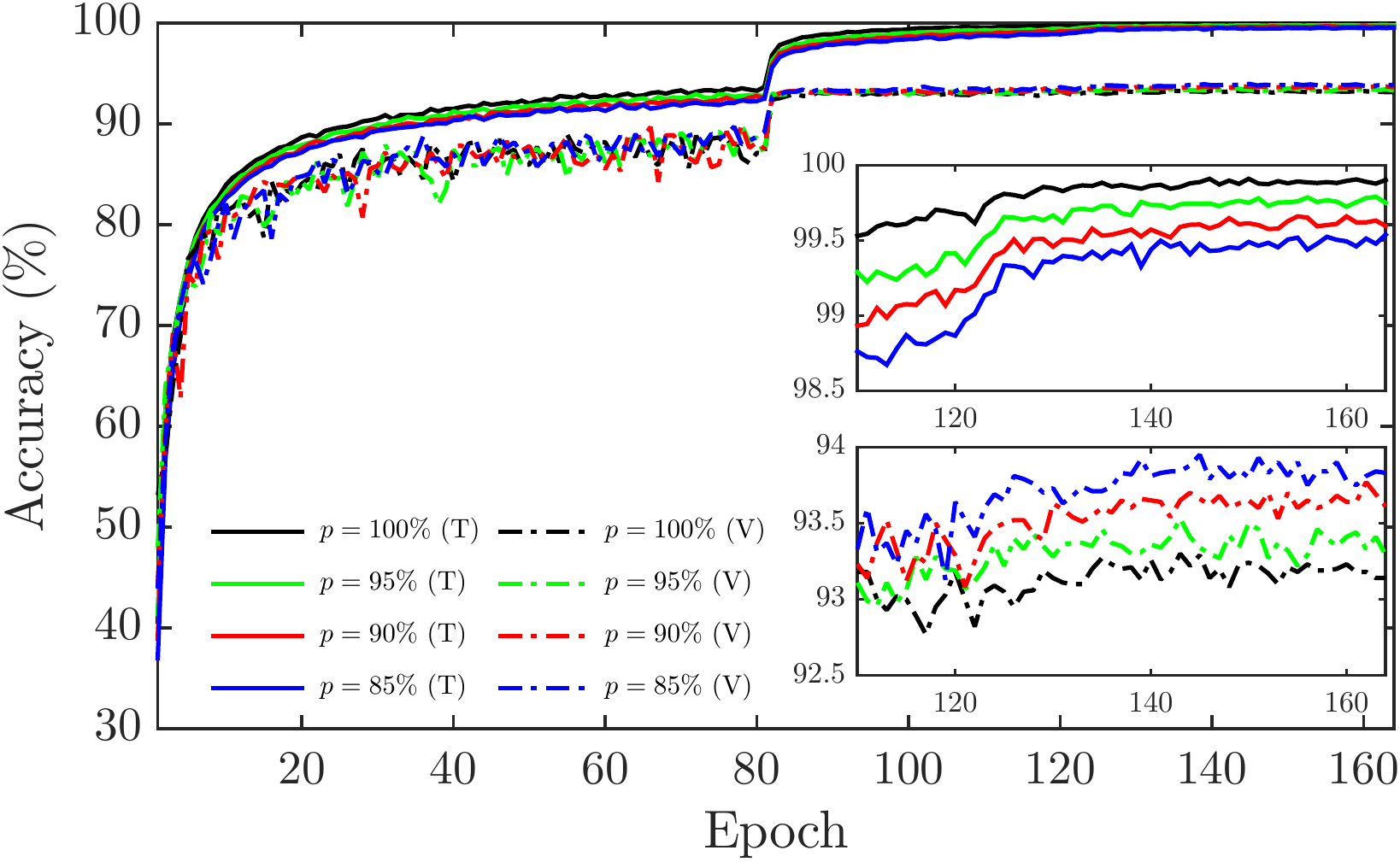}
        \caption{Accuracy for PreResNet-56.}
    \end{subfigure}

    \begin{subfigure}[t]{0.47\textwidth}
        \centering
        \includegraphics[width=0.98\textwidth]{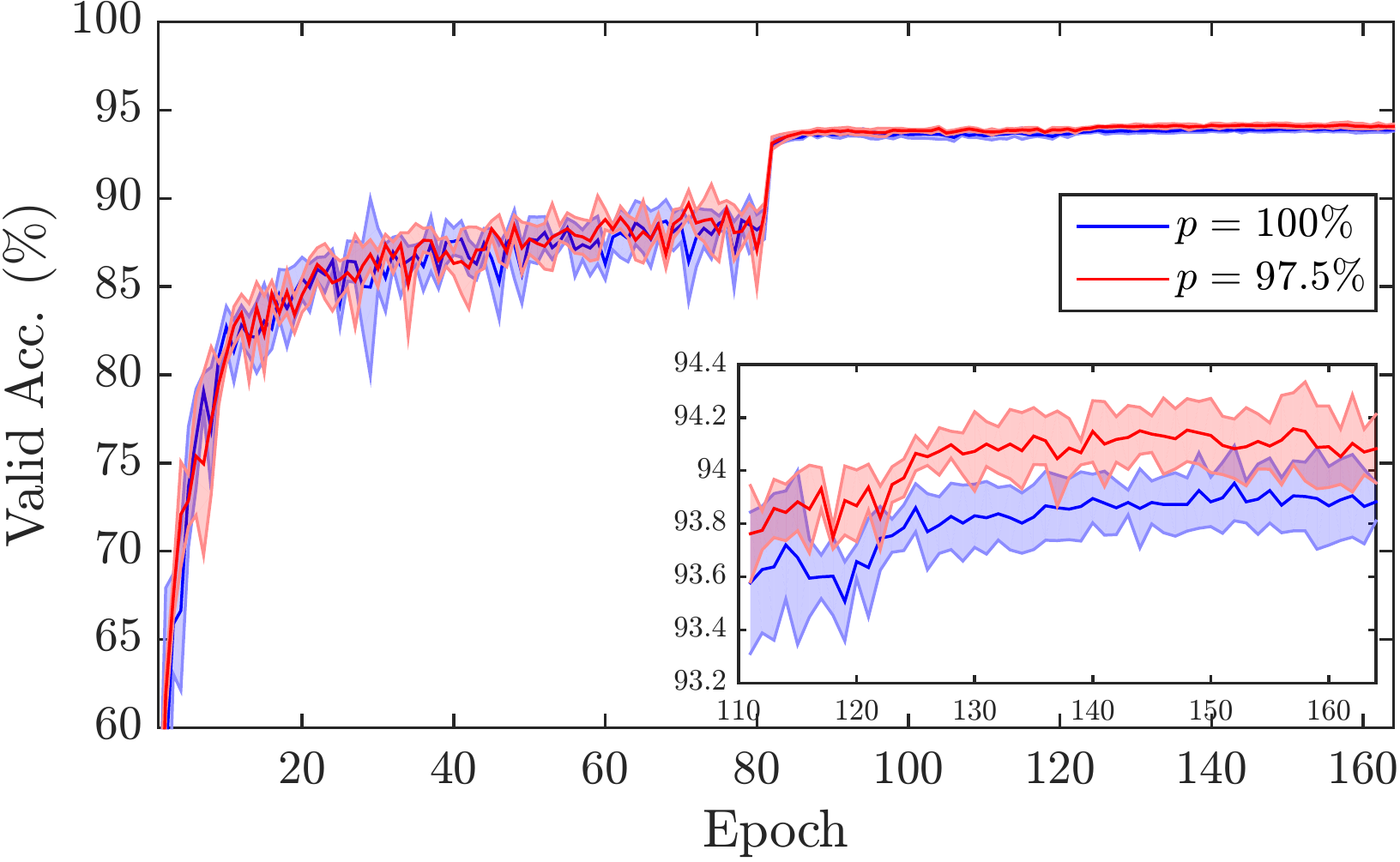}
        \caption{PreResNet-110 with $p=97.5\%$.}
    \end{subfigure}%
    ~ 
    \begin{subfigure}[t]{0.47\textwidth}
        \centering
        \includegraphics[width=0.98\textwidth]{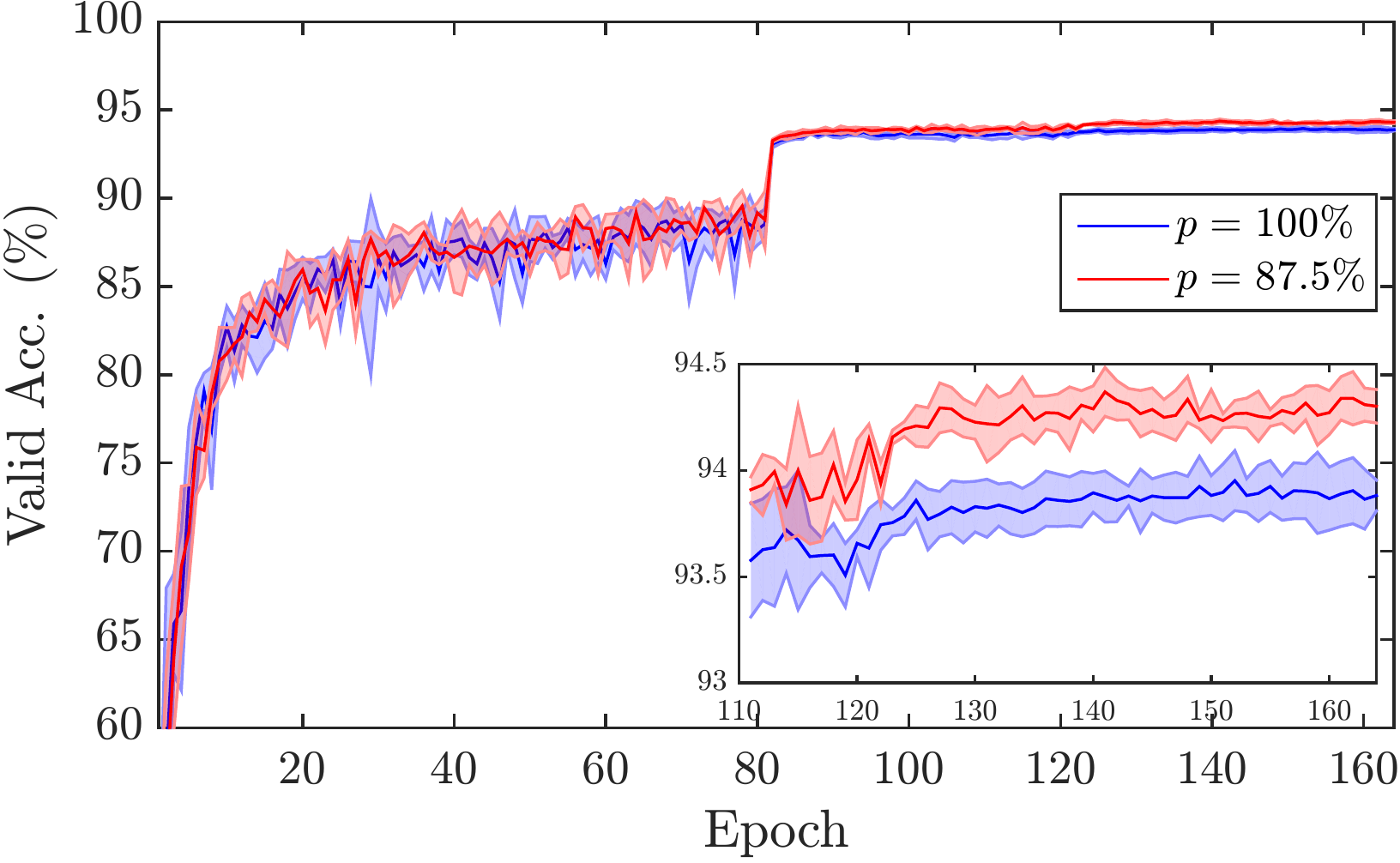}
        \caption{PreResNet-110 with $p=87.5\%$.}
    \end{subfigure}

    \begin{subfigure}[t]{0.47\textwidth}
        \centering
        \includegraphics[width=0.98\textwidth]{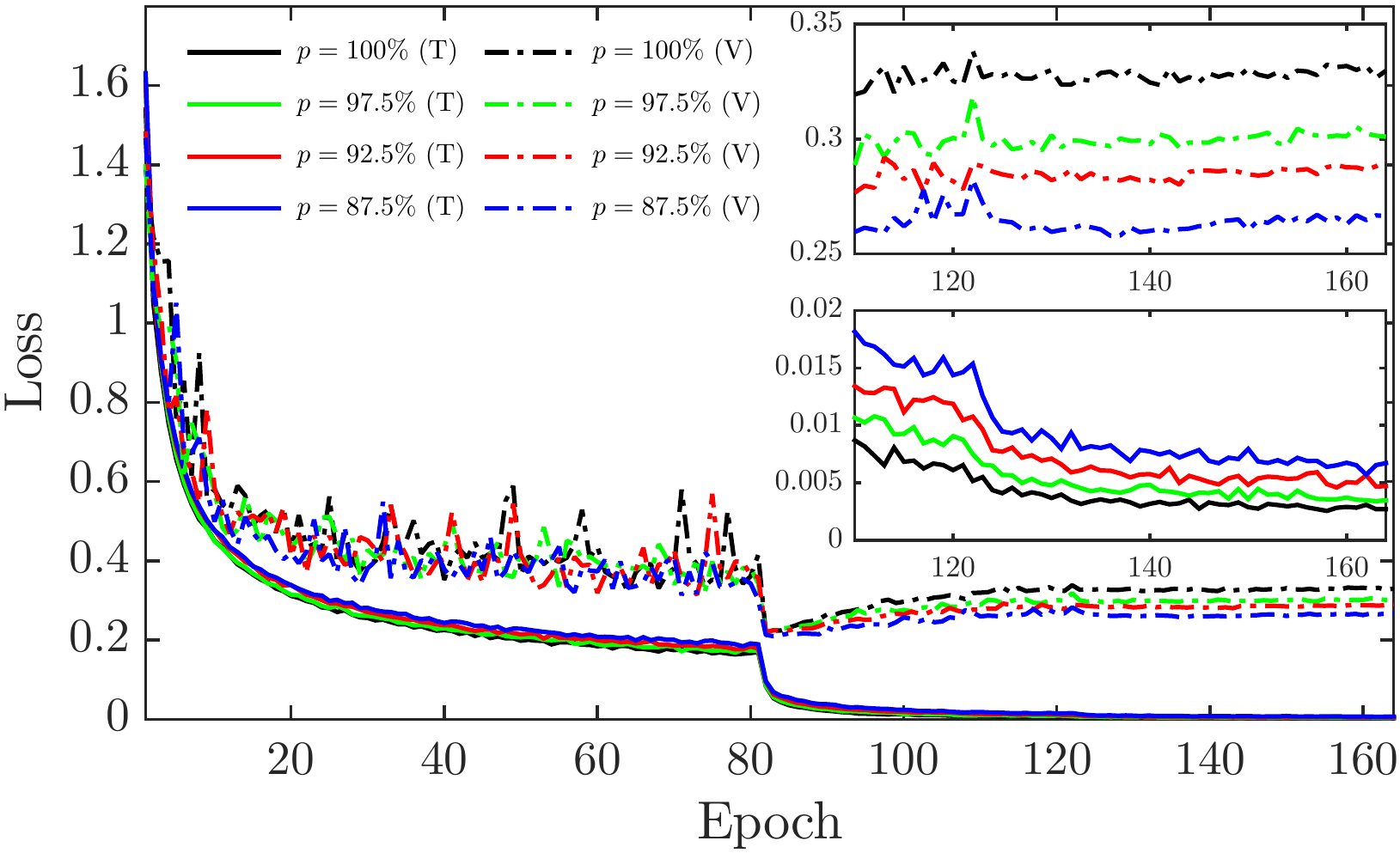}
        \caption{Loss value for PreResNet-110.}
    \end{subfigure}%
    ~ 
    \begin{subfigure}[t]{0.47\textwidth}
        \centering
        \includegraphics[width=0.98\textwidth]{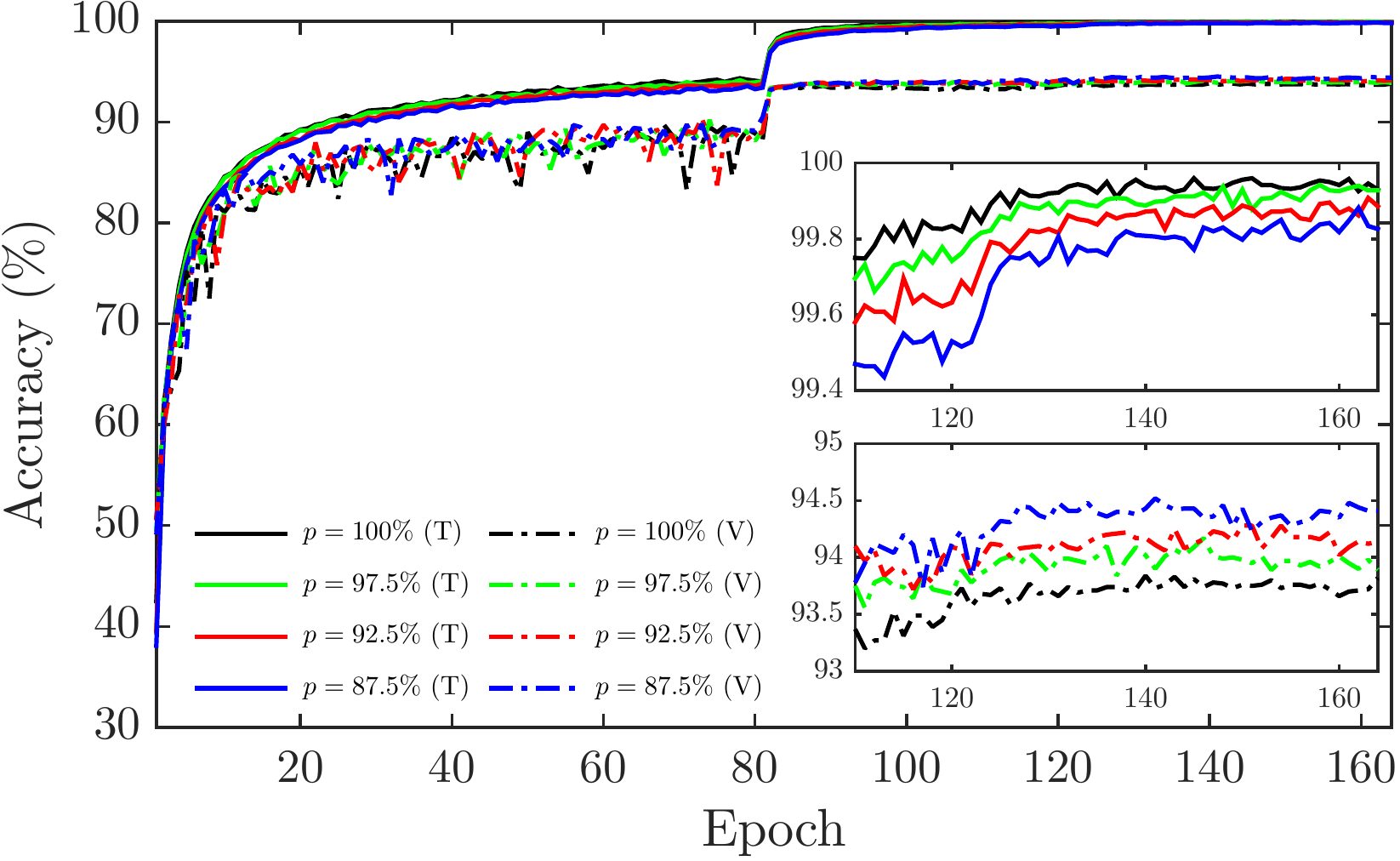}
        \caption{Accuracy for PreResNet-110.}
    \end{subfigure}
    \caption{Learning curves (mean $\pm$ variance) for PreResNet-56/110 with different dropout ratios, and learning curves for PreResNet-56/110 at training (T) and validation (V) during one simulation.}
    \label{Fig-R110-Sample-Path}
\end{figure*}

We also present the training and validation loss (accuracy) for PreResNet-20 of one simulation in \autoref{Fig-R20-Sample-Path}. By adding dropout layers, the gap between training and inference procedures is reduced for both the loss value and accuracy. In other words, we can trade some loss in training accuracy (or cross-entropy loss) for improvement in the generalization performance. Similar observations can be make for PreResNet-56/110 which validate our arguments, we refer the readers to \autoref{Fig-R110-Sample-Path} for more details.

\section{Conclusion and Future Work}
The method of (stochastic) modified equations is applied in this work to show that ResNet-like models with noise injection can be regarded as numerical discretizations of stochastic differential equations with multiplicative noise, which enables us to bridge a variety of stochastic training strategies with the optimal control of backward Kolmogorov's equations. Based on this connection, the training procedure of a plain ResNet can be naturally interpreted as the minimization of systems governed by transport equations. This finding brings us a novel perspective on the regularization effects of stochastic training techniques, \textit{e.g.}, the injected noise acts explicitly as a second-order artificial viscosity term in the backward Kolmogorov's equations which cuts down the number of poor local minima and forces the optimization algorithm toward optima that generalize well, which we hope to shed light on the design of more explainable and efficient stochastic training methods. To further characterize the regularization effects, a perspective of loss landscape is applied to a binary classification problem in dimension one, and experimental results for a real-world application support our arguments. 

Since stochastic training strategy incurs a trade-off between data fitting and model regularization, it will be  interesting to determine the optimal noise level associated with the depth of ResNet models. Moreover, in contrast to the aforementioned methods that regularize the label prediction, the rugged loss landscape could also be smoothed by convolution with a Gaussian kernel \cite{chaudhari2016entropy,chaudhari2017deep} or an elliptic smoothing operator \cite{osher2018laplacian}. The latter offers a regularization that can be achieved by simply modifying SGD algorithms during iteration. Furthermore,  one may combine both methods into a hybrid approach that forces optimization towards flat plateau regions to improve the generalization performance. We leave to future work more detailed analysis of these approaches and quantitative studies of tuning of regularization parameters.

\section*{Acknowledgments}
We acknowledge computing resources from Columbia University's Shared Research Computing Facility project, which is supported by NIH Research Facility Improvement Grant 1G20RR030893-01, and associated funds from the New York State Empire State Development, Division of Science Technology and Innovation (NYSTAR) Contract C090171, both awarded April 15, 2010. 

\bibliographystyle{siamplain}
\bibliography{references}

\end{document}